\documentclass[twocolumn,journal,10pt]{IEEEtran}
\usepackage{cite}
\usepackage[font=footnotesize]{subfig}
\usepackage{multirow}
\usepackage{ifpdf}
\ifCLASSINFOpdf
\usepackage[pdftex]{graphicx}
\else
\usepackage[dvips]{graphicx}
\fi
\usepackage{epstopdf}
\usepackage[cmex10]{amsmath}
\usepackage{algorithmic}
\usepackage{array}
\usepackage{multirow}
\usepackage{verbatim}
\usepackage[ruled,vlined]{algorithm2e}
\usepackage[font=footnotesize]{subfig}
\usepackage{amsmath,bm}
\usepackage{amssymb}
\usepackage{color}
\usepackage{subfig}
\usepackage{multirow}
\usepackage{float}
\usepackage{caption}
\begin{document}
	
	\title{Explainable Hierarchical Imitation Learning for Robotic Drink Pouring }

	\author{Dandan Zhang,~\IEEEmembership{Student Member, IEEE},  Qiang Li, Yu Zheng,~\IEEEmembership{Senior Member, IEEE}, Lei Wei,~\IEEEmembership{Member, IEEE}\\Dongsheng Zhang, Zhengyou Zhang,~\IEEEmembership{Fellow, IEEE} % <-this % stops a space
	\thanks{Dandan Zhang is with the Computing Department, Imperial College London, London, United Kingdom. E-mail: d.zhang17@imperial.ac.uk %This work is done when she is an intern at Tencent Robotics X. 
		\newline \indent Yu Zheng, Lei Wei, Dongsheng Zhang, and Zhengyou Zhang are with Tencent Robotics X, Shenzhen, Guangdong Province, China. E-mail: petezheng@tencent.com \newline \indent Qiang Li is with Neuroinformatics Group, Center for Cognitive Interaction Technology (CITEC), Bielefeld University, Bielefeld, Germany. }}

	\maketitle

\begin{abstract}
To accurately pour drinks into various containers is an essential skill for service robots. However, drink pouring is a dynamic process and difficult to model. 
Traditional deep imitation learning techniques for implementing autonomous robotic pouring have an inherent black-box effect and require a large amount of demonstration data for model training. To address these issues, an Explainable Hierarchical Imitation Learning (EHIL) method is proposed in this paper such that a robot can learn high-level general knowledge and execute low-level actions across multiple drink pouring scenarios. Moreover, with EHIL, a logical graph can be constructed for task execution, through which the decision-making process for action generation can be made explainable to users and the causes of failure can be traced out. Based on the logical graph, the framework is manipulable to achieve different targets while the adaptability to unseen scenarios can be achieved in an explainable manner.
A series of experiments have been conducted to verify the effectiveness of the proposed method. Results indicate that EHIL outperforms the traditional behavior cloning method in terms of success rate, adaptability, manipulability and explainability. 

\textit{Note to Practitioners---}
Pouring liquids is a common activity in people's daily lives and  all wet-lab industries. 	Drink pouring dynamic control is difficult to model, while the accurate perception of flow is challenging.  To enable the robot to learn under unknown dynamics via observing the human demonstration, deep imitation learning can be used. However, traditional deep imitation learning techniques have black-box effects, which means that the decision-making process is not transparent to ensure safety and is difficult to add trust to humans during task execution. 	To address the aforementioned limitations, an Explainable Hierarchical Imitation Learning (EHIL) method is proposed in this paper. In the EHIL, we decompose the pouring tasks into multiple hierarchies, by which a logical graph can be constructed. The proposed method can enable the robot to learn the knowledge of how and why to perform the task, rather than simply execute the task via traditional behavior cloning. In this way, explainability and safety can be ensured. Manipulability can be achieved by reconstructing the logical graph. 	The target of this research is to obtain pouring dynamics via the learning method and realize the precise and quick pouring of drink from the source containers to various targeted containers with reliable performance, adaptability, manipulability and explainability. 	  
\end{abstract}


% 1. Automatic Paremter Tuning
% 2. Bayesian Inference
% 3. Tranparency
% 4. Manipulability

\section{Introduction}
	
In recent years, service robots have been widely deployed in different environments such as houses, offices,
	hospitals \cite{rozo2016learning}.  They have been developed to perform a wide range of human activities such as cooking, cleaning, sweeping, personal care etc. Among the diverse activities mentioned above, pouring a specific amount of liquid into a container is one of the frequently performed tasks in food preparation, drink service or industrial environments, and is  significant for assisting people's daily life. Therefore, automatic drink pouring will be explored in this paper.

In most of the conventional robotized automation, robots are programmed to execute repetitive tasks  with high speed and high precision \cite{wang2018masd}. However, it is impractical to hard-code robots to conduct precise pouring in different scenarios with various types of drink and/or target containers, since it  involves dynamic processes that are difficult to model \cite{langsfeld2014incorporating,yamaguchi2015pouring}. Moreover, model-based approaches for programming the robot for task execution are  time-consuming and of low-efficiency.  Therefore, it is desirable to employ intelligent methodologies in  robotics and automation tasks to enable the quick deployment of service robot to assist humans' daily life \cite{oliff2020improving}.

\begin{figure}[tb]
	\centering
	\includegraphics[width=1\hsize]{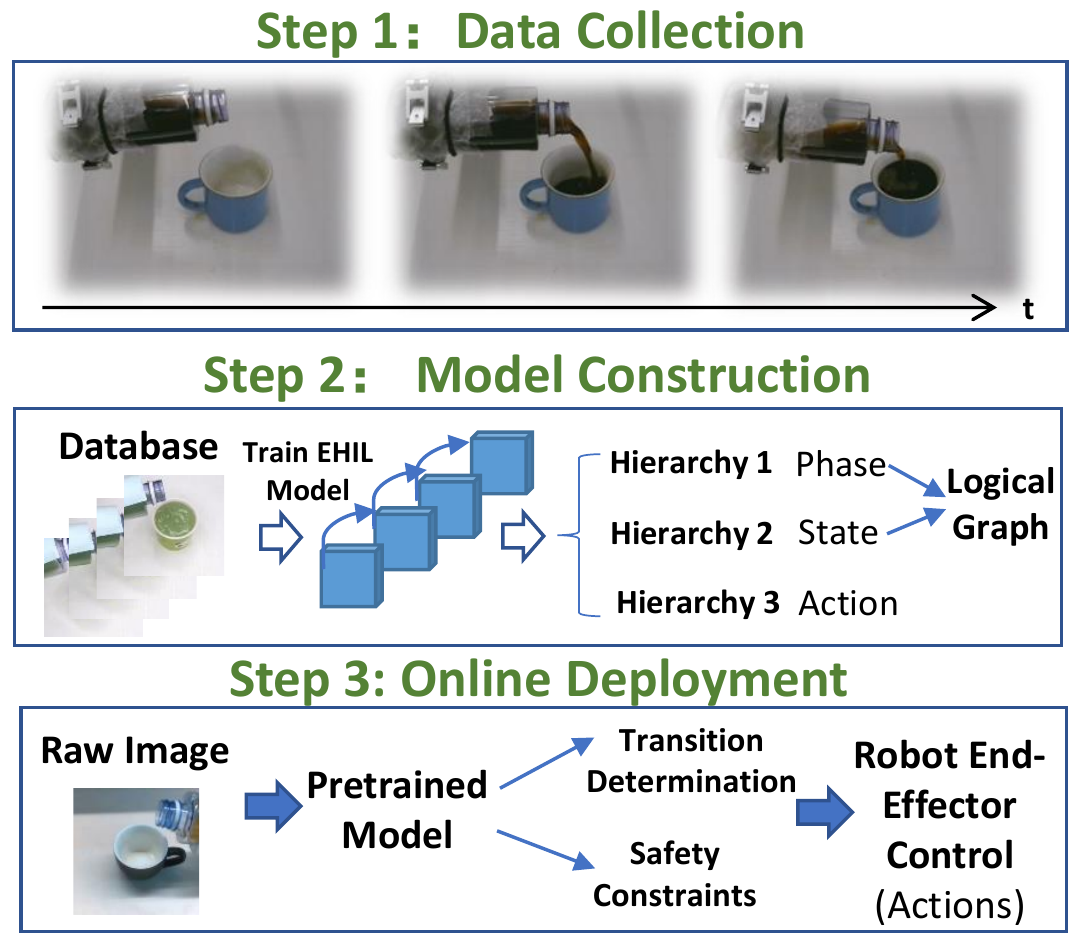}
	\captionsetup{font=footnotesize,labelsep=period}
	\caption{Workflow for the implementation of the EHIL method, including data collection, model construction and online deployment. }
	\label{Figure3-Workflow.pdf}
\end{figure}

Robot learning techniques, such as reinforcement learning and imitation learning, have been studied to enable the robots to acquire complex skills. 	
For example, a model-based reinforcement learning approach has been developed for liquid pouring task \cite{yamaguchi2016neural}, while  a model-free reinforcement learning approach based on Deep Deterministic Policy Gradients (DDPG) has been used for automatic drink pouring, which avoids the dynamic modeling \cite{do2018learning}.  However,  traditional  reinforcement learning methods require a time-consuming training process with an effective cost function to encode a specific robotic task, and is challenging to be implemented in practice when the trial-and-error process is expensive. 

%Moreover, the reality gap is difficult to be addressed, especially when the target robotic task involves complex dynamics.

Imitation learning \cite{asfour2008imitation}, as an alternative approach, can enable robots to pour liquids by providing them with expert's demonstrations \cite{Argall2009A,zhang2018deep,langsfeld2014incorporating,pan2017feedback}. It can offer a paradigm for robots to learn successful policies in fields where humans can easily demonstrate the desired behaviors but it is difficult to build a precise model for robot control. For example, a multimodal assembly skill decoding system  (MASD) has been developed to decode  the assembly skill performed in the human demonstration. It can be used to generate the robot programs, which promotes the flexibility of robotized automation \cite{wang2018masd}. A teaching-learning-collaboration (TLC) model has been developed for  robots to learn from human assembly demonstrations based on maximum entropy inverse reinforcement learning algorithm  to actively assist the human partner in the collaborative assembly task in shared working situations \cite{wang2018facilitating}. These methods provide a natural way to transfer knowledge from human to robot \cite{argall2009survey}, as well as avoid the tedious programming process. To this end, we aim to explore imitation learning  for automatic robotic pouring in this paper.  %with the main focus of implementing drink pouring in the service industry.

Deep imitation learning methods normally utilize a deep convolutional neural network (DCNN) with multiple layers to map pixels of visual observations to actions, which are trained via a supervised learning manner \cite{fu2019active}. It can enable the robot to distill states from raw sensory input to predict the next actions \cite{chen2019deep}. 
Previous works have investigated deep imitation learning through teleoperation \cite{chen2020supervised} or kinesthetic teaching \cite{zhang2018deep}. 
A large amount of data is required for model training, and the performance for task execution is significantly influenced when the pretrained model is applied to a new scene because of the domain shift.
For example, to achieve a good performance of reaching a single fixed object via end-to-end learning, around 200 demonstrations were used for model training \cite{zhang2018deep}. That is to say, a robot may struggle hard to accomplish a new task given limited demonstration data.  

Moreover, traditional deep imitation learning has inherent black-box effects; that is, the decision-making process is not transparent to users. The black box effects can be  acceptable in certain domains such as playing Atari games, but may impede the general acceptance of intelligent robots in tasks that involve humans or in critical applications. Due to the high cost of failure caused by wrong decisions, robots are expected to have the ability of reasoning failures in the case of executing wrong actions in the physical environment. To this end, we aim to develop intelligent robots that can generate explanations before the task execution, through which the user's trust in the robot's behavior can be built.
	
\begin{figure}[tb]
	\centering
	\includegraphics[width = 1\hsize]{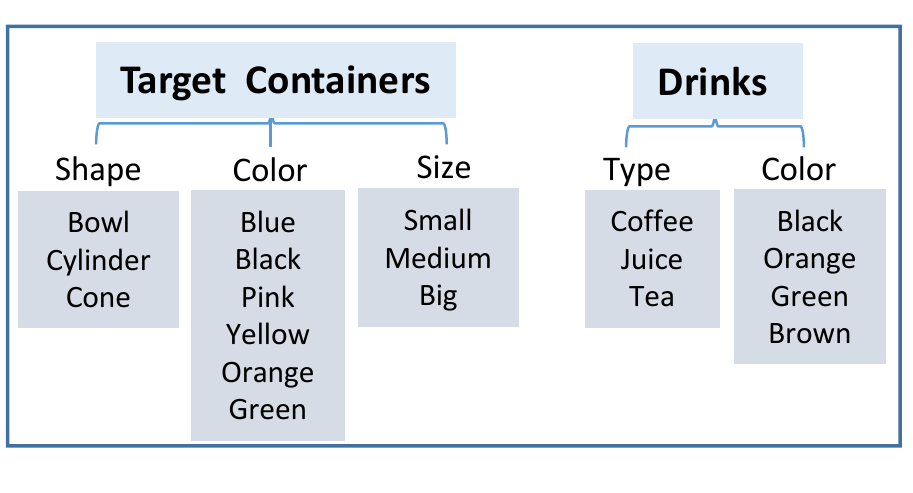}
	\captionsetup{font=footnotesize,labelsep=period}
	\caption{Overview of the database constructed for model training.  Different types of drink and different target containers  used for experiments.}
	\label{fig:Figure2-DrinkContainer.pdf}
\end{figure}
	
Locally weighted regression \cite{atkeson1991using},  gaussian process regression \cite{schneider2010robot,chen2020supervised}, decision trees \cite{konidaris2012robot}, RuleFit \cite{friedman2008predictive}, Bayesian inference \cite{zhang2018self}, K-nearest neighbors \cite{chen2016learning} and Dynamic Motion Primitive \cite{brandi2014generalizing} are popular approaches with explainable features. However, these methods cannot achieve comparable performance as deep learning based methods to pursue accurate prediction  \cite{yang2020enhancing,zhang2020data,zhang2020automatic}. 
Therefore, the balance between explainability and model performance should be considered. 
To address the aforementioned issues while ensuring the  performance for task execution, this paper presents an Explainable Hierarchical Imitation Learning (EHIL) method, which is featured by i) a hierarchical structure and ii) an explainable logical graph.

Hierarchical learning \cite{parr1998reinforcement,fox2019multi} has been used to divide a task into several sub-tasks for robot learning and proved to have generalization across various tasks with sequential features~\cite{xu2018neural}. Hierarchical deep convolutional neural network has been proven to have high performance with respect to accuracy for automatic detection and classification \cite{xie2019automatic}.
Formulating imitation learning in a hierarchical structure can break down the drink pouring process into several sub-processes and allow a robot to make good use of the information included in the demonstration. 
After discovering the general concepts from demonstrated data, the robot can learn to perform drink pouring task in new scenes which have not been demonstrated before. 
	
In traditional neural network based approaches, though they have demonstrated strong capabilities of automatic noise removal, the relations between the images fed to a neural network and the generated actions are unknown. Then, it is almost impossible for users to understand how an action is determined by the network based on a given image. Therefore, explainable learning algorithms which aim to interpret the decision-making process deserve to be explored \cite{sanneman2020trust}. 
To address the issue of lacking transparency and add trustworthiness for users when using the model, a logical graph is developed for the proposed EHIL method. 
The logical graph can leverage intuitive and useful explanations due to its expressive nature. 
With the logical graph, manipulability for the task execution can be ensured. 
This means that the proposed method can be flexible to deal with different scenarios when a different target amount of drink is required to be poured into the target container. 
The logical graph constructed based on the first and second hierarchies can be reorganized if a new task is required to be performed while the general	knowledge can be reused. 
In this way, humans can manipulate the model without retraining the model from scratch and pouring drink in new scenarios can be performed without recollecting demonstration data.	
	
Moreover, the proposed method has the adaptability to some extent. Although models are trained with demonstrations of topping up empty containers, without retraining with new demonstrations, they can be applied to new scenarios. For example, the algorithm can be applied to the scene where a target container is not empty at the initial state or additional drink or ices are added to the target container during the drink pouring process. This cannot be achieved in most of the existing deep learning based algorithms. For example, an RNN (Recurrent Neural Network) based MPC (Model Predictive Control) controller has been employed for liquid pouring \cite{chen2019accurate}. A recurrent LSTM CNN method has been utilized for visual closed-loop control for liquids pouring \cite{schenck2017visual}. The performance of the execution of a specific task can be ensured with RNN, which is good at dealing with sequential decision-making tasks. However, the mechanism of memory may bring negative effects when the initial status of the target container is changed or the pouring process is interrupted by humans.

To summary, the main advantages of the proposed EHIL method include i) explainability, ii) manipulability, and iii) adaptability.	The EHIL method decomposes the learning process into three hierarchies, where the first hierarchy divides a pouring task into several phases, the second hierarchy monitors the liquid state in the target container, and the third hierarchy determines the action for task execution. 
A logical graph is constructed based on the first and second hierarchies and enables users to conduct safety checks before  task execution or reason the failures after task execution, making the decision-making process  intuitive and interpretable. 
In other words, the EHIL method can generate action commands together with the corresponding state values to ensure that a robot performs a task based on the desired logical graph leading to the success of the task. In case that some unexpected failures occur, the causes of failure can be traced out and explained. The framework is manipulable to enable the robot to achieve different pouring targets by reconstructing the logical graph, through which the robots can perform the pouring task in new scenes without model retraining.

The rest of this paper is structured as follows. The details of the EHIL method are presented in Section II, followed by experiments in Section III to verify its performance, adaptability, manipulability and explainability.
Section IV concludes this paper with discussions on the current limitations and potential improvements of the method.

\section{Methodology}

\subsection{Overview}

The workflow for the implementation of the  EHIL method is shown in Fig.~\ref{Figure3-Workflow.pdf}, which consists of three steps, namely data collection, model construction, and online deployment. In Step 1, demonstration data is collected via teleoperation, which forms a database. Following that, a hierarchical model is trained and a logical graph is constructed in Step 2.  As for online deployment in Step 3, the pretrained model with state and phase transition determination and safety constraints are used to determine the real-time action for robot end-effector control, which leads to the implementation of automatic drink pouring.
Technical details are illustrated as follows.

\begin{figure}[tb]
	\centering
	\includegraphics[width=1\hsize]{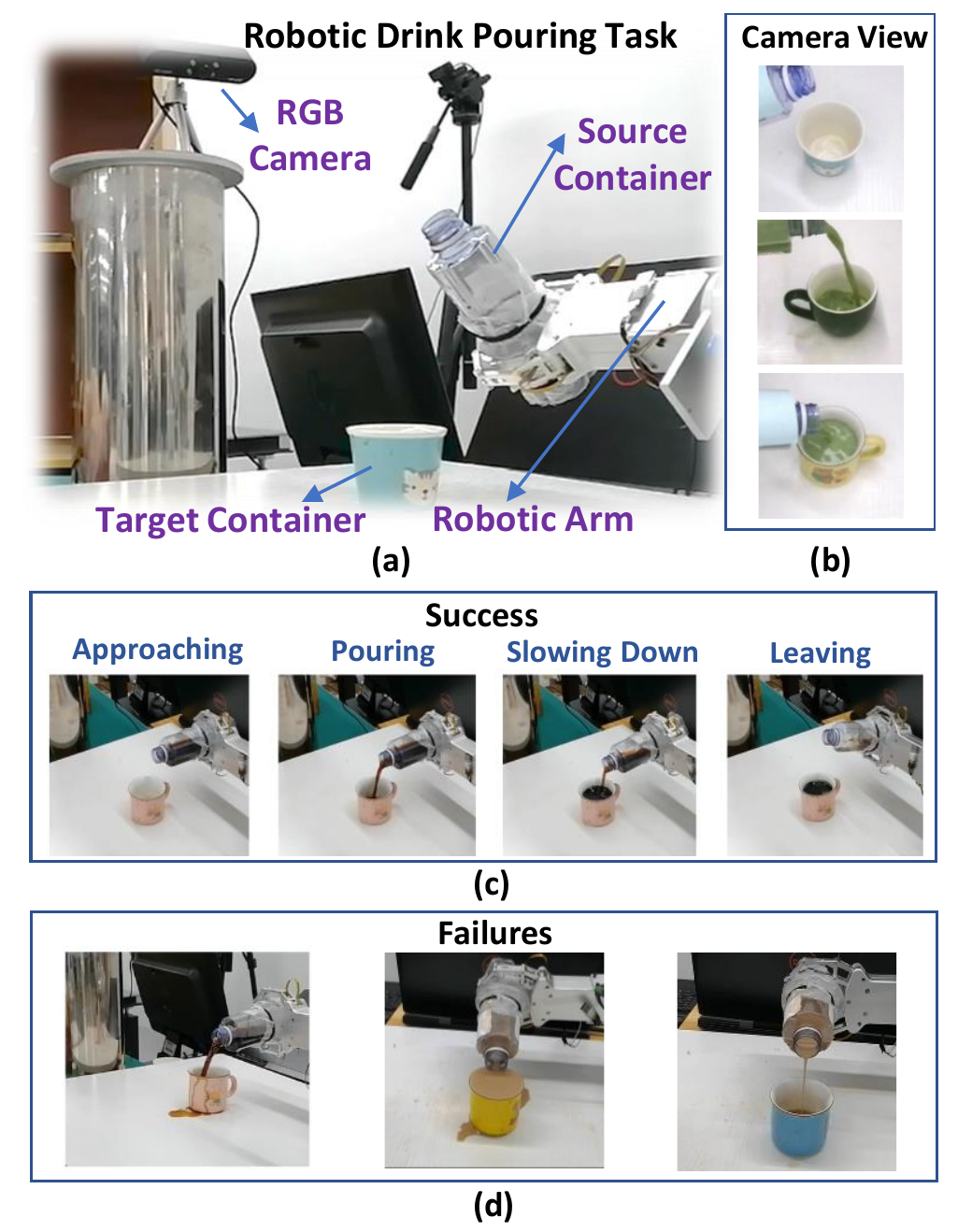}
	\captionsetup{font=footnotesize,labelsep=period}
	\caption{Overview of the setup and scenes of robotic drink pouring. (a) Setup for data collection and experimental validation. (b) Examples of images obtained by an RGB camera. (c) Four phases of a successful drink pouring task. (d) Typical failures including drink spilling and not filling the target container with the target amount of drink.}
	\label{fig:Figure1-SystemOverview-New.pdf}
\end{figure}

\subsection{Data Collection}

Fig. \ref{fig:Figure1-SystemOverview-New.pdf}a shows our setup for conducting robotic drink pouring task, which includes a UR-16E robotic arm, an RGB camera, a source container held by a customized two-finger gripper, and a target container on the table. 
Demonstrations are performed via teleoperation, where an in-house master manipulator is maneuvered by a human operator to remotely control the robotic arm.

The demonstration data comprises the images of drink pouring acquired by an RGB camera with a resolution of 640$\times$480 and the  velocities of the gripper obtained through forward kinematics of the robotic arm. The velocity value collected at each time step $t$ is a 6-D vector, which is consist of the 3-D linear velocity and the 3-D angular velocity. 
To pour drink automatically, the rolling angle of its wrist joint is controlled,	through which the tilt angle of the source container can be controlled directly.

Both visual and kinematic data are collected at a frequency of 25 Hz and synchronized to form a series of ground-truth trajectories consisting of image-action pairs.
Fig. \ref{fig:Figure1-SystemOverview-New.pdf}b shows some examples of the camera view, which captures the source and target containers, the drink flowing out of the source container into the target container, and the drink in the target container.

\begin{table}[tb]
	\captionsetup{font=footnotesize,labelsep=period}
	\caption{\textsc{Different Drink Pouring Scenes for Data Collection}}
	\label{tab:n1}
	\centering
	\begin{tabular}{c|c|c|c|c|c}
		\hline
		\multirow{2}{*}{\textbf{Scene}} &	\multicolumn{3}{c|}{\textbf{Target Container}} &  \multicolumn{2}{c}{\textbf{Drink}} \\\cline{2-6}
		& \textbf{Shape}   & \textbf{Size} & \textbf{ Color} & \textbf{Type} & \textbf{Color}   \\\hline
		\textbf{1}  & Bowl &   Small & Black &  Coffee & Brown \\
		\textbf{2}  & Cone & Medium & Green & Tea &  Green \\
		\textbf{3}  & Cylinder &   Big & Pink & Coffee & Brown   \\
		\textbf{4}  & Cylinder & Big & Yellow   & Tea &  Green\\
		\textbf{5}  &  Cylinder & Big& Orange & Coffee & Brown \\			
		\textbf{6}  & Cone & Medium & Pink &  Coffee & Brown   \\		
		\textbf{7}  & Cylinder & Big  & Pink & Coffee & Brown \\
		\textbf{8}  & Bowl   & Small &  Pink &  Coffee & Brown \\
		\textbf{9}  &  Cone & Medium & Blue  & Tea & Orange\\
		\textbf{10}  & Cone &  Medium & Green & Tea& Orange \\	 	
		
		\textbf{11}  & Cylinder & Big & Blue &  Juice & Orange\\
		\textbf{12}  & Cone & Medium & Yellow & Coffee &  Black \\
		\textbf{13}  & Bowl &   Small & Pink &  Coffee & Black   \\
		\textbf{14}  & Cone & Medium & Blue   & Juice & Orange \\
		\textbf{15}  &  Cylinder & Big& Orange & Tea& Green \\			
		\textbf{16}  & Cone & Medium & Green & Coffee & Black \\		
		\textbf{17}  & Cylinder & Big  & Pink & Coffee & Black \\
		\textbf{18}  & Bowl   & Small & Black &  Tea& Green \\
		\textbf{19}  &  Bowl  & Small & Pink  &Juice & Orange  \\
		\textbf{20}  & Cone &  Medium & Yellow & Tea& Green  \\	 		
		\hline
	\end{tabular}
\end{table}

We define different pouring scenes to be the cases where either target containers or drinks are different. Target containers have different shapes, colors and sizes, while drinks have different types and colors. We consider the size of the target container to be big if its maximum capacity is greater than 150 ml or small if it is less than 100 ml and medium in between. 

For model training, we constructed a drink pouring demonstration dataset with 80 trials in 20 different scenes, as listed in Table \ref{tab:n1}. In each trial, we let the robot pour certain drink into an empty target container and filling it up to at least 90\% without spilling and label every phase of a successful task, as depicted in Fig.~\ref{fig:Figure1-SystemOverview-New.pdf}c.  Several typical scenes of failures are shown in Fig.~\ref{fig:Figure1-SystemOverview-New.pdf}d, which indicate i) the drinks spill due to excessive pouring action, ii) the drinks overflow due to wrong action determination, and iii) the container is not filled enough.

\subsection{Hierarchical Model Construction}

\begin{figure*}[tb]
	\centering
	\includegraphics[width = 1\hsize]{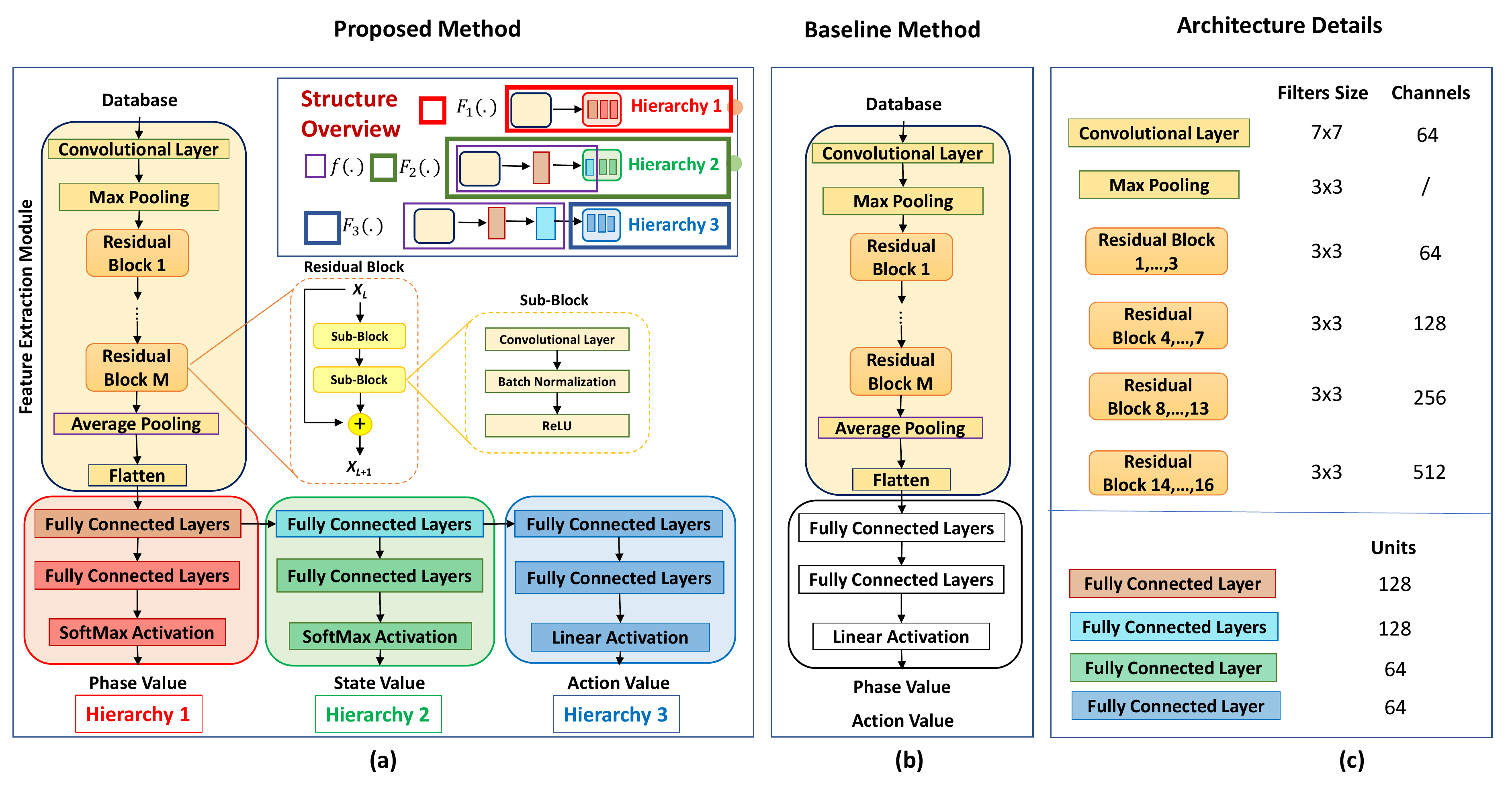}
	\captionsetup{font=footnotesize,labelsep=period}
	\caption{ Overview of (a) the proposed Explainable Hierarchical Imitation Learning (EHIL) method, (b) the baseline model for comparative study, and (c) the details for the model construction of the EHIL method.}
	\label{fig:Network}
\end{figure*}

\begin{figure*}[tb]
	\centering
	\includegraphics[width = 1\hsize]{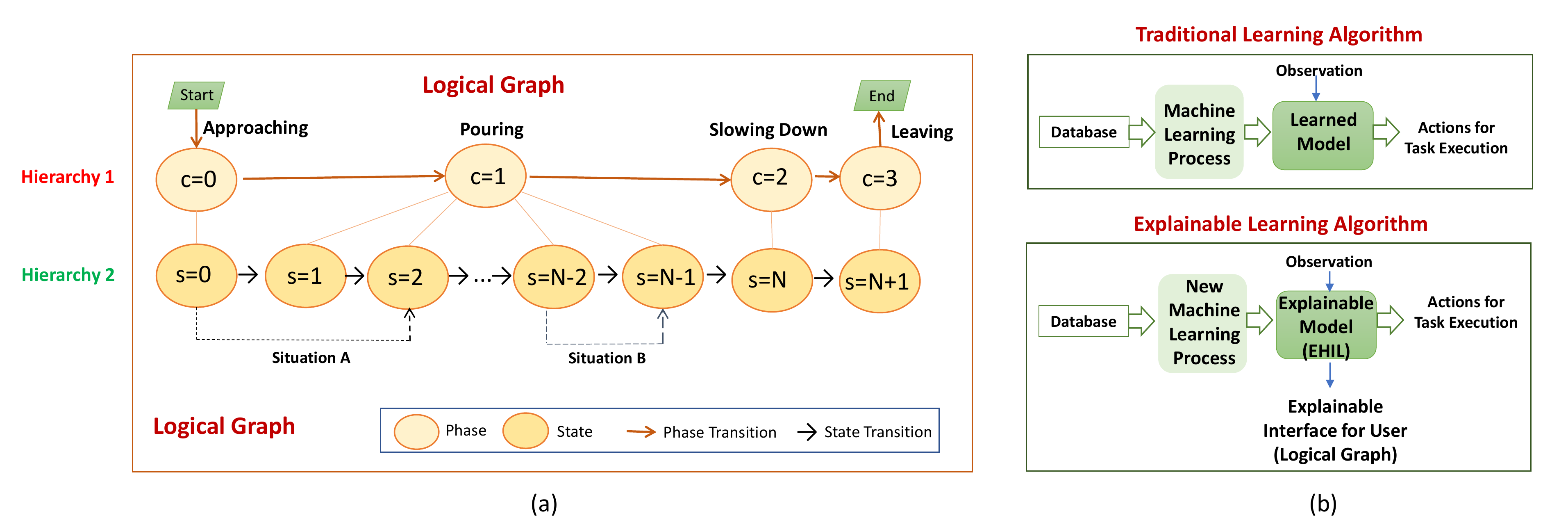}
	\captionsetup{font=footnotesize,labelsep=period}
	\caption{ Overview of the logical graph and the concept of explainable learning. (a) Illustration of the logical graph constructed by the first and second hierarchies. (b) Comparison of the traditional learning algorithm and the explainable learning algorithm.}
	\label{fig:XAI-Frame.pdf}
\end{figure*}

\subsubsection{Hierarchical Model Architecture}

Deep neural networks have achieved remarkable success in image classification tasks and the effectiveness of imitation learning has been proved \cite{zhang2018deep}. However, it is time-consuming to train deep neural networks and may suffer gradient vanishing problems. Deep residual neural networks can accelerate the training process and have better performance than the traditional neural network \cite{he2016deep}. Therefore, we employ a deep residual neural network for our task.

Fig.~\ref{fig:Network}a gives an overview of the hierarchical structure of EHIL, which consists of three hierarchies denoted by $F_i(.)$, $i=1,2,3$. 
The three hierarchies have a common module for image feature extraction, as depicted by the yellow block in Fig.~\ref{fig:Network}a. 
It consists of a convolutional layer followed by a max pooling layer, multiple residual blocks, and an average pooling layer, while the generated 2-D feature array of an input image is reshaped to a 1-D feature vector after passing through the last flatten layer.
The residual blocks are constructed by two convolutional layers with shortcut connection and element-wise addition.   A stride of 2 is performed when the shortcuts go across feature maps of two sizes.
	
Fig.~\ref{fig:Network}b shows the architecture of a baseline method used for the comparative study in this paper. It employs a ResNet-34 architecture and is trained in an end-to-end manner to determine the desired pouring angle from raw pixels~\cite{he2016deep}, which can be known as the traditional behavior coloning approach. The baseline method has the same number of residual blocks, convolutional layers and fully connected layers as the model for the third hierarchy of EHIL.
	
The details of the construction of residual blocks can be referred to \cite{he2016deep}. 33 convolutional layers with different size and number of filters are used to construct the feature extraction module in this paper, while the details for the construction of the model are shown in Fig.~\ref{fig:Network}c.  The number of convolutional layers and the residual blocks can be customized to meet the requirements of different applications.  	Batch normalization and Relu activation are connected after each convolutional layer  before the next convolutional layer \cite{he2016deep}.

	After feature extraction, the 1-D feature vector is fed into the subsequent fully-connected layers and activation functions for phase and state classifications and action regression. 
	More details of the model training for the three hierarchies are provided as follows.

	\subsubsection{Hierarchical Model Training}
	The first hierarchy is aimed at pouring phase determination, which enables the robot to reason about the phase of  pouring.
	We divide a drink pouring task into four phases with labels $c=0,1,2,3$, as depicted in Fig.~\ref{fig:Figure1-SystemOverview-New.pdf}c:
	\begin{enumerate}
		\item \textit{Approaching} ($c=0$): The robot is tilting the source container but the drink has not come out yet;
		\item \textit{Pouring} ($c=1$): The robot keeps tilting the source container and the drink is flowing into the target container;
		\item \textit{Slowing down} ($c=2$): The robot is retreating to slow down pouring while the target container is nearly full;
		\item \textit{Leaving} ($c=3$): The target container is full and the robot restores the source container to its original pose with no more drink flowing into the target container.
	\end{enumerate}
	The first hierarchy can be deemed as a common sense for drink pouring and is invariant in different pouring scenes. 
	Let $\mathcal{O}_1$ denote the dataset of the image data obtained from the demonstration database with corresponding phase labels annotated by humans. It is used to train model $F_1(.)$ for the first hierarchy.
	
	The second hierarchy is intended for pouring state determination such that the robot can estimate the amount of drink that has been poured into the target container.
	The neural network structure is the same as the first hierarchy except that more fully-connected layers are added to improve the classification accuracy, since the state determination process is more complex compared with phase determination. 
	To reduce the model training time for $F_2(.)$, the parameters of model $F_1(.)$ can be used to initialize and expedite the training of model $F_2(.)$, which preserves important feature extraction knowledge compared with random initialization.
	We manually segment the image sequences of drink pouring in each robot demonstration into $N$ states and label them as $s=0,1,2,\dotsc,N,N+1$.
	Assume that the total amount of drink to be poured is $V_{all}$, which equals the size of the target container in the training data. Then, the resolution for liquid state estimation is $k=V_{all}/N$, while the target container is empty initially, which is labeled as $s=0$. 
	Let $\mathcal{O}_2$ denote the labeled image dataset for training model $F_2(.)$ for the second hierarchy.
	
	Let $\mathcal{L}_i({\Theta}_i, \mathcal{O}_i)$ ($i=1,2$) denote the loss function (cost function) for training model $F_i(.)$, which is written as
	\begin{equation}
		\label{A}
		\mathcal{L}_{i}({\Theta}_{i}, \mathcal{O}_{i})=-\sum_{j}^{N_{i}} t_{j} \log \left(\frac{e^{s_{j}}}{\sum_{k}^{N_{i}} e^{s_{k}}}\right)
	\end{equation}
	where ${\Theta}_i$ represents the model parameters, $\mathcal{O}_i$ is the labeled dataset, $t_j$ and $s_j$ represents the ground truth data and the score predicted by the neural network model for each class $j$ in $N_i$ classes, respectively.

	The third hierarchy is used to determine the robot action by mapping image array of visual observations to the angular velocity of the robot's end-effector. 
	Since the action generation has close relationship with the drink pouring state determination, the model of the third hierarchy shares a significant part of the network with the second hierarchy.
	Let $f(.)$ denote the shared structure of the second and third hierarchies, which acts as a sub-model for feature embedding before training model $F_3(.)$. 
	The model $F_3(.)$ is constructed by multiple fully-connected layers, followed by a linear activation. 
	Once model $F_2(.)$ is trained, the parameters of $f(.)$ are inherited from $F_2(.)$ and directly applied for feature embedding before training $F_3(.)$. 
	The relations among $F_i(.),i=1,2,3$ and $f(.)$ are illustrated in Fig.~\ref{fig:Network}a.

	Suppose that $\mathcal{O}_3$ represents the dataset of images associated with corresponding velocities of the end-effector in robot demonstrations. 
	$\mathcal{O}^{'}_3 = f(\mathcal{O}_3)$ contains the feature vectors obtained by feeding the image data of $\mathcal{O}_3$ to $f(.)$, while the corresponding labels remain unchanged. 
	Then, the embedded features are used for training the parameters of model $F_3(.)$.
	The loss function for the third hierarchy is taken to be the mean squared error and can be written as
	\begin{equation}
		\label{cc}
		\mathcal{L}_3({\Theta}_3, \mathcal{O}^{'}_3)=\frac{\sum_{j}^{n}\left(t_j-s_j\right)^{2}}{n}
	\end{equation}
	where $t_j$ and $s_j$ are the ground truth and the value predicted by the neural network model during regression, respectively, and $n$ is the number of samples used for training.

\begin{algorithm}[tb]
	\SetAlgoLined
	\textbf{Input:}  $\mathcal{O}_i,~i=1,2,3$ \\
	%Initial parameters $\widehat{{\Theta}}_i$ for $F^{h}_{i}(.)(i=1,2)$;\\
	
	Initialize parameters ${\Theta}_i$ for $F_{i}(.),~i=1,2,3$ \\
	\For{i=1,2}
	{	\While{Training $F_{i}({\Theta}_i,\mathcal{O}_{i})$}{
			Sample a batch of data from the dataset \\	
			Compute the loss function $\mathcal{L}_{i}\left({\Theta}_i, \mathcal{O}_{i}\right)$ \\
			Set ${\Theta}_i \leftarrow {\Theta}_i - \alpha \nabla_{{\Theta}_i} \mathcal{L}_{i}({\Theta}_i, \mathcal{O}_{i})$ \\
		}	
	}
	Extract $f(.)$ after training $F_2(.)$ \\
	\While{Training $F_{3}({\Theta}_3,\mathcal{O}_{3})$}{
		Sample a batch of data from the dataset \\	
		$O^{'}_{3} \leftarrow f(\mathcal{O}_{3})$ \\
		Compute the loss function $\mathcal{L}_{3}({\Theta}_3, \mathcal{O}^{'}_{3})$ \\	
		Set ${\Theta}_3 \leftarrow {\Theta}_3 - \alpha \nabla_{{\Theta}_3} \mathcal{L}_{3}({\Theta}_3, \mathcal{O}^{'}_{3})$ \\
	}	
	\textbf{Output:} $F_{i}(.),~i=1,2,3$
	\caption{Model Training of the EHIL Method}
\end{algorithm}

The training of the hierarchical model for the proposed EHIL method is described in \textbf{Algorithm 1}.
	Optimization is performed using Adam \cite{kingma2014adam} with a learning rate of ${\alpha} = 0.0001$ and a batch size of 20.

	\subsubsection{Logical Graph}
	
	A logical graph can be constructed with the outputs of the first and second hierarchies to reveal the rules governing the robot's behavior in terms of logical status and transitions, as illustrated in Fig.~\ref{fig:XAI-Frame.pdf}a.
	For any typical drink pouring task that is to top up an empty container, like those in the collected demonstrations, the logical graph consists of a sequence of state transitions from $s=0$ over $s=1,\dotsc,N$ to $s=N+1$, where $s=0$, $s\in[1,N)$, $s=N$, and $s=N+1$ correspond to the ``approaching" ($c=0$), ``pouring" ($c=1$), ``slowing down" ($c=2$), and ``leaving" ($c=3$) phases, respectively. 
	Formulating the	high-level decision-making process as a human-readable logical graph has several advantages including
	\begin{itemize}
		\item Making the decision-making process of the robot explainable to users can enhance their trust in task execution;
		\item The causes of failure can be traced out in the logical graph.	 
	\end{itemize}
	
	The explainable learning algorithm aims to interpret the decision-making process~\cite{sanneman2020trust} and a comparison with the traditional learning algorithm is shown in Fig.~\ref{fig:XAI-Frame.pdf}b.	We focus on the techniques to extract information	from the trained neural network models for the purpose of 	generating explanations and justifications that help users to 	understand the robot's behavior. The hierarchical structure proposed in this paper can be regarded as an explainable model, while the logical graph can serve as an explainable interface for users.

	\subsection{Online Deployment}
	For the online deployment, the image of a pouring scene is captured by an RGB camera and fed into the pretrained models to determine the pouring phase and state as well as the desired robot action.
	To ensure a stable phase or state transition, the probability of a determined phase or state is thresholded to decide whether we trust the current determination or retain the previous one. For service robots, they are expected to perform tasks intelligently while preserving safety. 
	Therefore, safety constraints are added to the determined robot action to ensure that the action to be executed is within a reasonable range.
	
	\subsubsection{Transition Determination}
	
	Let $c({t-1})$ and $c({t})$ be the pouring phases determined by EHIL at times $t-1$ and $t$, respectively, where $c({t}) = 0,1,2,3$. 
	Let ${c'}(t)$ be the output of model $F_1(.)$ at time $t$, which is an intermediate result before deciding the actual phase label $c(t)$. The goal of transition determination is to find out $c(t)=\mathcal{B}({c'}(t))$, which determines the pouring phase by rectifying the prediction of $F_1(.)$ in case of noisy outputs.
	
	Since a drink pouring task is a sequential process, ${c'}(t)$ should be within the range $[c({t-1}),c({t-1}) + 1]$. 
	Suppose that the probability of the output of $F_1(.)$ at time $t$ is $\mathcal{P}(c = c^{*})$, where $c^{*} = 0,1,2,3$.
	Then, ${c'}(t)$ is taken to be the phase with the highest probability, i.e.,
	\begin{equation}
		\label{c}
		{c'}(t) = \textit{argmax}\{\mathcal{P}(c=c({t-1})),...,\mathcal{P}(c=3)\}
	\end{equation}
	
	When the probability of ${c'}(t)$, namely $\mathcal{P}(c={c'}(t))$, is slightly greater than $\mathcal{P}(c=c({t-1}))$, we assume that the phase remains unchanged. 
	We accept ${c'}(t)$ as $c({t})$ only when $\mathcal{P}(c={c'}(t))$ is sufficiently leading, which implies that the estimation of ${c'}(t)$ has been stable and can be well trusted. 
	Hence, $\mathcal{B}({c'}_t)$ can be written as
	\begin{equation}
		\mathcal{B}({c'}(t)) = \left\{
		\begin{array}{ll}
			c(t-1) & \mathrm{if}~{\Delta}\mathcal{P}_c < \lambda \\
			{c'}(t) & \mathrm{if}~{\Delta}\mathcal{P}_c \geq \lambda
		\end{array} \right. 
	\end{equation}
	where ${\Delta}\mathcal{P}_c={\mathcal{P}(c={c'}(t)) - \mathcal{P}(c=c({t-1}))}$ and ${\lambda} \in [0,0.5)$ is a threshold, which can be tuned online and is taken to be 0.2 in this paper.
	Finally, the pouring phase at time $t$ is obtained by $c(t)=\mathcal{B}({c'}(t))$.

	Similarly, at time $t$, we take the current state to be $s(t)=\mathcal{B}({s'}(t))$, where as
	\begin{equation}
		\label{eq:state_transition}
		\mathcal{B}({s'}(t)) = \left\{
		\begin{array}{ll}
			s({t-1}) & \mathrm{if}~ {\Delta}\mathcal{P}_s< \lambda \\
			{s'}(t) & \mathrm{if}~{\Delta}\mathcal{P}_s \geq \lambda
		\end{array} \right. 
	\end{equation}
	where $s(t-1)$ is the previous pouring state determined by EHIL, ${s'}(t) = \textit{argmax}\{\mathcal{P}(s=s({t-1})),...,\mathcal{P}(s=N)\}$ is the state determined by model $F_2(.)$ at time $t$, and ${\Delta}\mathcal{P}_s={\mathcal{P}(s={s'}(t)) - \mathcal{P}(s=s({t-1}))}$.
	
	\subsubsection{Safety Constraints}
	Based on a statistical analysis of the collected robot demonstrations, we can obtain the minimum and maximum required angular velocities, denoted by $\underline{\omega}_s$ and $\bar{\omega}_s$, respectively, of the robot's end-effector for every pouring state $s$.
	Let $\omega'(t)$ be the end-effector's angular velocity determined by the model $F_3(.)$ at time $t$.
	Then, we regulate the desired angular velocity for controlling the robot as $\omega(t) = \mathcal{M}(\omega'(t),s)$, where
	\begin{equation}
		\mathcal{M}(\omega'(t),s) =\left\{
		\begin{array}{ll}
			\underline{\omega}_s & \mathrm{if}~\omega'(t)\leq \underline{\omega}_s \\
			\omega'(t) & \mathrm{if}~\underline{\omega}_s<\omega'(t)<\bar{\omega}_s \\
			\bar{\omega}_s & \mathrm{if}~\omega'(t) \geq \bar{\omega}_s
		\end{array} \right. 
	\end{equation}
	
	Similarly, from collected robot demonstrations, we can obtain the minimum and maximum required tilting angles of the robot's end-effector for every pouring phase $c$, which are denoted by $\underline{\theta}_c$ and $\bar{\theta}_c$, respectively.
	Let $\theta'(t) = \theta(t-1)+\omega(t) \delta t$, where $\theta(t-1)$ is the previous angle of the robot's end-effector and $\delta t$ is the timestep for integration. 
	Then, the desired angle for commanding the robot is $\theta(t) = \mathcal{M}(\theta'(t),c)$, where
	\begin{equation}
		\mathcal{M}(\theta'(t),c) =\left\{
		\begin{array}{ll}
			\underline{\theta}_c & \mathrm{if}~\theta'(t)\leq \underline{\theta}_c \\
			\theta'(t) &  \mathrm{if}~\underline{\theta}_c<\theta'(t)<\bar{\theta}_c \\
			\bar{\theta}_c & \mathrm{if}~\theta'(t) \geq \bar{\theta}_c
		\end{array} \right. 
	\end{equation}

\begin{comment}
	\begin{algorithm}[tb]
\SetAlgoLined
\textbf{Input:} $o_t$ (image array of current visual observation)\\
$\omega_0$, $\omega_{e}$ (approaching and leaving velocities) \\
$F_{i}(.)$, $i=1,2,3$ (pretrained models) \\
$[\underline{\theta}_c, \bar{\theta}_c]$, $c=0,1,2,3$ (tilting angle domain) \\
$[\underline{\omega}_s, \bar{\omega}_s]$, $s=1,2,\dotsc,N$ (angular velocity domain) \\	
$\theta_0$, $t = 0$  (initial robot pose and time) \\
\While{Pouring task has not done}
{

$c \leftarrow \mathcal{B}(F_{1}(o_{t})) $  \\
\If{$c = 0$}
{
$\omega(t) \leftarrow \omega_0$
}

\If{$c = 1$ or $c = 2$}
{
$s \leftarrow \mathcal{B}(F_{2}(o_{t})) $  \\
$\omega(t) \leftarrow \mathcal{M}(F_{3}(o_t),s)$  \\
}
\If{$c = 3$}
{
$\omega(t) \leftarrow \omega_{e}$
}
$\theta(t) \leftarrow \mathcal{M}(\theta(t-1)+\omega(t){\delta}t,c)$  \\
$t = t + 1$
} 
\textbf{Output:}  $\omega(t)$ and/or $\theta(t)$ for robot control
\caption{Online Deployment of the EHIL Method}
\end{algorithm}
\end{comment}

 \begin{algorithm}
	\SetAlgoLined
	\textbf{Input:}
	$s_{\mathrm{goal}}$ (goal state) \\
	$\omega_0$, $\omega_{N+1}$ (approaching and leaving velocities) \\
	$F_{i}(.)$, $i=1,2,3$ (pretrained models) \\
	$[\underline{\theta}_c, \bar{\theta}_c]$, $c=0,1,2,3$ (tilting angle domain) \\
	$[\underline{\omega}_s, \bar{\omega}_s]$, $s=1,2,\dotsc,N$ (angular velocity domain) \\	
	$\theta_0$, $t = 0$  (initial robot pose and time) \\
	\While{Pouring task has not done}
	{
		$o_t \leftarrow$ current observation  \\
		$c \leftarrow \mathcal{B}(F_{1}(o_{t})) $  \\
		\If{$c = 0$}
		{
			$\omega(t) \leftarrow \omega_0$
		}
		
		\If{$c = 1$ or $c = 2$}
		{
			$s \leftarrow \mathcal{B}(F_{2}(o_{t})) $  \\
			\eIf{$s < s_{\mathrm{goal}}$}
			{
				$\omega(t) \leftarrow \mathcal{M}(F_{3}(o_t),s)$  \\	
			}
			{
				$\omega(t) \leftarrow \mathcal{M}(F_{3}(o_t),N)$  \\
			}
		}
		\If{$c = 3$}
		{
			$\omega(t) \leftarrow \omega_{N+1}$
		}
		$\theta(t) \leftarrow \mathcal{M}(\theta(t-1)+\omega(t){\delta}t,c)$  \\
		$t = t + 1$
	}
	
	\textbf{Output:}  $\omega(t)$ and $\theta(t)$  \\
	\caption{Online Deployment of the EHIL Method}
\end{algorithm}

	\subsubsection{Online Control}

	\textbf{Algorithm 2} describes the online control of a robot to perform the drink pouring task with the proposed EHIL method. 
	Suppose $F_{i}(.)(i=1,2,3)$ have been obtained by \textbf{Algorithm 1}, the weights of the hierarchical neural network models and parameters obtained for state transition determination as well as safety constraints via statistic analysis are then used for online deployment.
	
	Model $F_1(.)$ is used to assist the determination of the phase based on the current observation. While the determined phase is ``approaching'' ($c=0$) or ``leaving'' ($c=3$), the robot turns or recovers its end-effector at a constant speed $\omega_0$ or $\omega_{e}$, which are obtained via linear regression based on the demonstration data. For the other two phases, the pouring state is further determined by model $F_2(.)$ and the desired angular velocity for controlling the robot is determined by model $F_3(.)$.

Although models are trained with demonstrations of topping up empty containers, without retraining with new demonstrations, they can be applied to other situations. We define this capability as adaptability. For example, when the target container is not empty, model $F_2(.)$ can determine the right state, say state $s=2$ as depicted by Situation A in Fig.~\ref{fig:XAI-Frame.pdf}a. Then, state $s=1$ is skipped and the state transitions of the pouring task become $s=0,2,\dotsc,N,N+1$. Therefore, less drink is poured and the reduction or the number of skipped states reflects the amount of drink initially in the target container.  In the meantime, human can add additional drink to the target container when the robot is pouring, as depicted by Situation B in Fig.~\ref{fig:XAI-Frame.pdf}a. More detailed definitions of these two situation and the experimental verification of the adaptability of the proposed method are illustrated in Section III-B. 
	
Moreover, the proposed model also has manipulability to some extend, which means that the model can be used when a different amount of drink needs to be poured into a target container. In this case,  we intentionally skip the last (few) state(s) in phase $c=1$. This can be achieved by setting a final goal state $s_{\mathrm{goal}}$ in \textbf{Algorithm 2}, where $s_{\mathrm{goal}} \in [1,N)$.
	In this way, the amount of drink to be poured into the target container can be manipulated, which by contrast cannot be realized by traditional end-to-end imitation learning. The experimental verification of the adaptability of the proposed method are illustrated in Section III-C.

	\section{Experiments}
	
	We have conducted five sets of experiments to validate the proposed method. The results are reported and discussed in this section.
	The experiment setup is shown in Fig. \ref{fig:Figure1-SystemOverview-New.pdf}a. 
	Besides the drink pouring scenes included in robot demonstrations (see Table~\ref{tab:n1}), a number of new scenes with unseen drinks and target containers are used to verify the applicability of our method, as listed in Table~\ref{tab:new situation}.

	With the experiments, we tend to answer three questions:
	\begin{itemize}
		\item How is the performance of the proposed EHIL method compared with the traditional behavior cloning method to perform drink pouring in different scenes? (the first set of experiments)
		
		\item Can the algorithm demonstrate its adaptability by performing new drink pouring tasks with various amount of drink without recollecting training data or retraining models? (the second set of experiments)
		
		\item	Can the algorithm
	be manipulated to perform new pouring tasks with different final goals? (the third set of experiments)
		
		\item Can explainability enhance the user's trust and help trace the causes of failures? (last three sets of experiments)
	\end{itemize}
%	Please see our \href{https://www.youtube.com/watch?v=qm9TnU05Bx8&feature=youtu.be}{supplemental video}.

	\begin{table}[tb]
		\centering
		\captionsetup{font=footnotesize,labelsep=period}
		\caption{\textsc{New Drink Pouring Scenes for Evaluation}}
		\label{tab:new situation}
		\begin{tabular}{c|c|c|c|c|c}
			\hline
			\multirow{2}{*}{\textbf{Scene}} &	\multicolumn{3}{c|}{\textbf{Target Container}} &  \multicolumn{2}{c}{\textbf{Drink}} \\\cline{2-6}
			& \textbf{Shape}   & \textbf{Size} & \textbf{ Color} & \textbf{Type} & \textbf{Color}   \\\hline
			\textbf{21}   &  Bowl  & Small  & Pink  & Tea & Green \\
			\textbf{22}  & Cylinder & Big  & Blue &  Coffee& Black \\	
			\textbf{23}  & Cone & Medium  & White & Tea & Green \\
			\textbf{24}   &  Bowl  & Small  & Green  & Juice & Orange \\   
			\textbf{25}  & Bowl  & Small & Blue & Tea & Green \\
			\textbf{26}   &  Bowl  & Small  & Grey  & Juice & Orange \\
			\textbf{27}  & Cylinder & Big  & Green &  Coffee& Black \\	
			\textbf{28}   &  Cylinder  & Big  & Blue  & Coffee & Brown \\  
			\textbf{29}   &  Cylinder & Big   & Orange  & Coffee& Black\\   
			\textbf{30}   &  Cone & Medium  & White  & Coffee& Black\\  
			\textbf{31}   &  Bowl & Small & Grey  &  Coffee& Brown \\   
			\textbf{32}   &  Cone & Medium  & Blue  & Coffee& Brown \\
			\textbf{33}   &  Bowl & Small  & Green  & Juice & Orange 	
			\\	
			\textbf{34}   &  Cylinder & Big  & Yellow  & Tea & Orange 	
\\ 		
			\hline
		\end{tabular}
	\end{table}

\subsection{Performance Evaluation}
	
For performance evaluation, we provide offline analysis results to demonstrate the properties of the pre-trained neural network models. Following that, parameter tuning is conducted for enhancing the effectiveness of the logical graph based on probability analysis. Based on the pre-trained neural network model and fine-tuned parameters for online deployment, physical experiments were conducted based on the UR16E robot. We compare the proposed method with the traditional behavioral cloning method (baseline) for performance evaluation.

	\subsubsection{Offline Analysis}
Offline analysis is conducted to evaluate the  neural network property based on the classification accuracy for phase and state estimation in hierarchies 1 and 2, respectively, as well as the regression error for action generation in hierarchy 3 of EHIL.

 80\% of the data is used for training, while 20\% of the data is used for testing. The training and testing accuracy for the first and second hierarchies, along with the Mean-Square-Error (MSE) for training and testing of the third hierarchy are  shown in Table \ref{tab:Results-Explain1}.

\begin{table}[!htb]
	\centering
	\caption{Offline analysis results for the performance evaluation of the model for EHIL.}
	\label{tab:Results-Explain1}
	\begin{tabular}{c|cc}
		\hline
		\textbf{Hierarchy}   & \textbf{Training}  & \textbf{Testing} \\\hline
		\textbf{Hierarchy1} & 98.68\% &98.14\%  \\
		\textbf{Hierarchy2} & 93.99\% & 91.09\%  \\
		\textbf{Hierarchy3}  &    0.0036 & 0.0013 \\
		\hline
	\end{tabular}

\end{table}

The accuracy of context awareness is high (more than 98\%) for both the training and testing dataset, which means that the robot can execute the task by following the general workflow of pouring. The state estimation accuracy is 93.09\% and 91.09\% for the training and testing dataset, respectively. The errors may occur due to the inaccurate labeling between two consecutive states. The state estimation accuracy should be further enhanced. However, the overall performance of the model is good enough for task execution since the transition determination and safety constraints for online deployment can help compensate for the errors.

\subsubsection{Parameter Tuning}

\begin{figure*}[tb]
	\centering
	\includegraphics[width = 1\hsize]{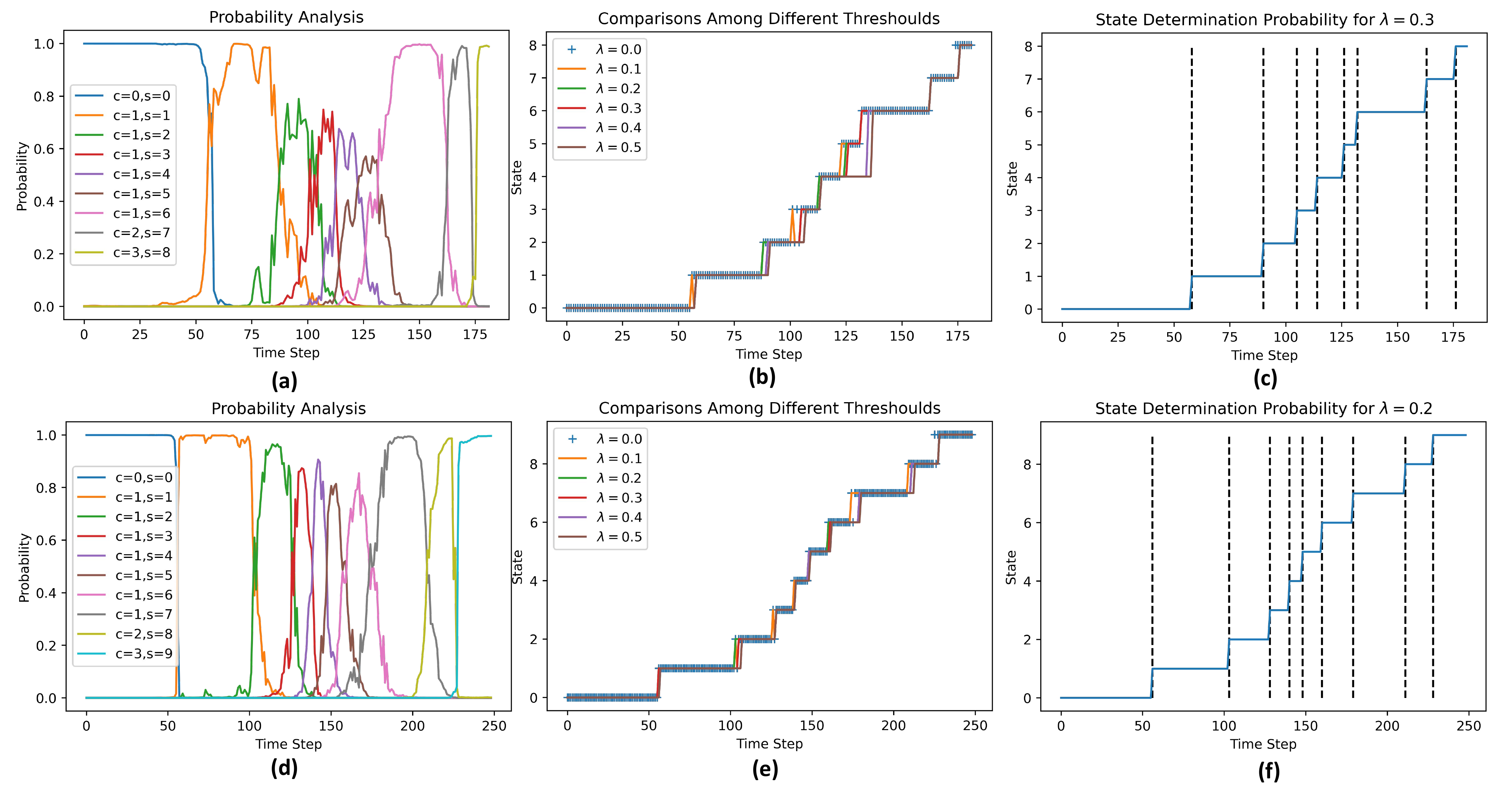}
	\captionsetup{font=footnotesize,labelsep=period}
	\caption{Analysis of the logical graph based on probability for improvement via parameter tuning. (a) Probability analysis for Scene 14. (b) Comparison among different threshold used for state transition estimation for Scene 14. (c) Final state estimation and transition process of Scene 14 by using the most suitable parameter. (d) Probability analysis for Scene 15. (e) Comparisons among different threshold used for state transition estimation for Scene 15. (f) Final state estimation and transition process of Scene 15 by using the most suitable parameter.}
	\label{fig:TuneParameters}
\end{figure*}
Results of the parameter tuning for the logical graph based on probability analysis are shown in Fig. \ref{fig:TuneParameters}. It can be seen that when ${\lambda}<=0.2$, the state estimation is fluctuated between state 2 and state 3 based on the neural network prediction. If ${\lambda}>0.3$, state 5 is skipped (see Fig.~\ref{fig:TuneParameters}b). Therefore, ${\lambda}=0.3$ is chosen as the threshold in this case, the state estimation and transition process is visualized in Fig.~\ref{fig:TuneParameters}c. Similarly, ${\lambda}=0.2$ is selected as the threshold value for conducting drink pouring task of Scene 15. The probability analysis, comparisons among different threshold values and the final state transition determination results for Scene 15 are demonstrated in Fig.~\ref{fig:TuneParameters}d--f.

\subsubsection{Success Rate}
	
\begin{figure}[tb]
	\centering
	\includegraphics[width = 1\hsize]{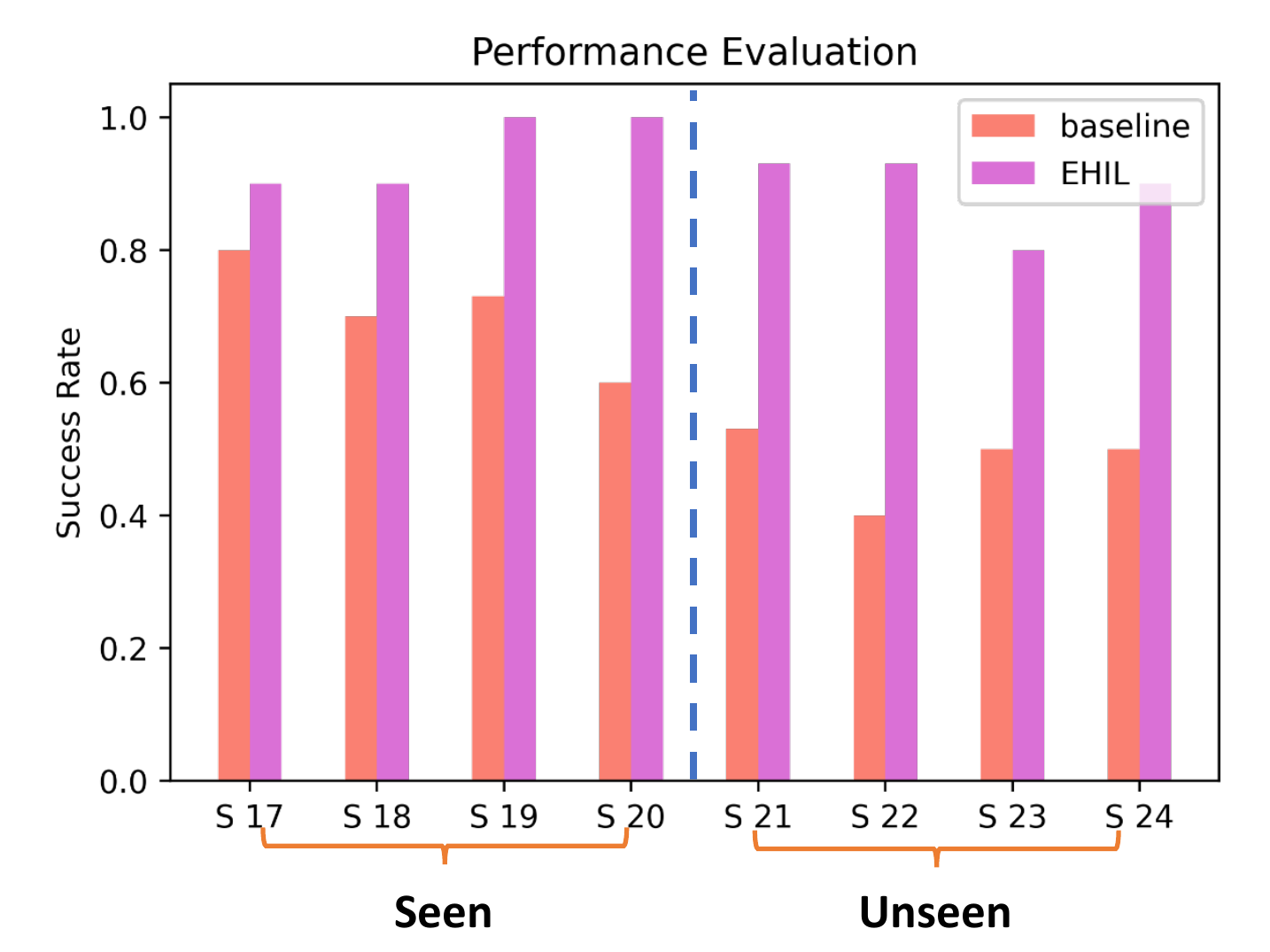}
	\captionsetup{font=footnotesize,labelsep=period}
	\caption{Success rates of the baseline method and the proposed EHIL method.} 
	\label{fig:performance}
\end{figure}
% naturalness; smoothness

The first set of physical experiments is intended to verify the applicability of the proposed EHIL method compared with the traditional behavior cloning method as the baseline method.

Eight different scenes, including four seen (Scenes 17--20 in Table~\ref{tab:n1}) and four unseen (Scenes 21--24 in Table~\ref{tab:new situation}) ones, are selected for tests. Ten to fifteen trials are conducted for each scene, while 100 trials are conducted in total. The performance of either algorithm is evaluated by its success rate in trials.	Here, successful pouring is defined to have no spilling during the whole process and fill at least 90\% of the target container. 

Fig.~\ref{fig:performance} shows the comparison of the success rates of the two methods, while the details are illustrated in Table \ref{tab:Results0}.	The overall success rate of EHIL for performing all the trials is 0.92, which is improved by 56\% from 0.59 of the baseline method. More specifically, in the seen and unseen scenes, the success rates of EHIL are 0.96 and 0.88, versus 0.70 and 0.48 of the baseline method, respectively. The success rate of EHIL for performing unseen pouring tasks is nearly twice the one of the baseline method. Based on these results, it can be concluded that the proposed EHIL method outperforms the baseline method.

\begin{table}[tb]
	\centering
	\captionsetup{font=footnotesize,labelsep=period}
	\caption{\textsc{Success Rates of the Baseline and Our EHIL Method}}
	\label{tab:Results0}
	\begin{tabular}{c|cc}
		\hline
		\textbf{Scene}  & 	\textbf{Baseline} & 	\textbf{EHIL}    \\\hline
		\textbf{17-20} (Seen)  &  35/50      &  48/50  \\
		\textbf{21-24} (Unseen)  &   24/50   & 44/50 \\
		\textbf{17-24} (Overall)&   59/100   & 92/100 \\	
		\hline
	\end{tabular}
\end{table}

	\subsection{Adaptability}

	%	Fig.~\ref{fig:Network}b,
	The second set of experiments shows that, to some extent, the proposed method has adaptability to new tasks of pouring various amount of drink to the target container that have not been demonstrated before.
	As depicted in Fig.~\ref{fig:XAI-Frame.pdf}a, we consider two situations as detailed below:
	\begin{itemize}
		\item \textit{Situation A}: In the case that the target container is not empty initially, the state transition of the pouring task may skip one or several states at the beginning of the pouring process and become $s=0,k,\dotsc,N,N+1$, where $1 < k < N$.
		\item \textit{Situation B}: Users can add additional drinks or ice cubes during the pouring task such that the state transition may skip states in the course of the process and become $s=0,\dotsc,k_1,(\dotsc),k_2,\dotsc,N,N+1$, where $1<k_1<k_2<N$.
	\end{itemize}
	In both situations, less drink will be poured and the number of skipped states reflects the reduced amount of drink poured into the target container.	
		\begin{table}[tb]
		\centering
		\captionsetup{font=footnotesize,labelsep=period}
		\caption{\textsc{Results of Adaptability Validation}}
		\label{tab:Manipulability11}
		\begin{tabular}{c|c|c|c | c|c|cc|c}
			\hline
			\multicolumn{4}{c|}{\textbf{\textit{Situation A}}} & 
			\multicolumn{5}{c}{\textbf{\textit{Situation B}}}\\\hline
			\textbf{Case} & \textbf{Scene}& $k$ & \textbf{SR} & 	\textbf{Case} & \textbf{Scene}& $k_1$ & $k_2$ & \textbf{SR}\\\hline
			\textbf{C1}	&	17	&  5  & 1.00 & 	\textbf{C5}	&	7	   & 2  & 3  & 0.60\\
			\textbf{C2}	&	18  & 2  & 0.90 & 	\textbf{C6}	&	19	   & 3  & 5  & 0.70\\ 	
			\textbf{C3}	&	19  & 3  & 0.90& 	\textbf{C7}	&	31	   & 2  & 3  & 0.80\\
			\textbf{C4}	&	27	& 2  & 0.90 & \textbf{C8}	&	33	   & 3  & 4 & 0.70\\		 	
			\hline
		\end{tabular}
	\end{table}

\begin{figure}[tb]
	\centering
	\includegraphics[width = 1\hsize]{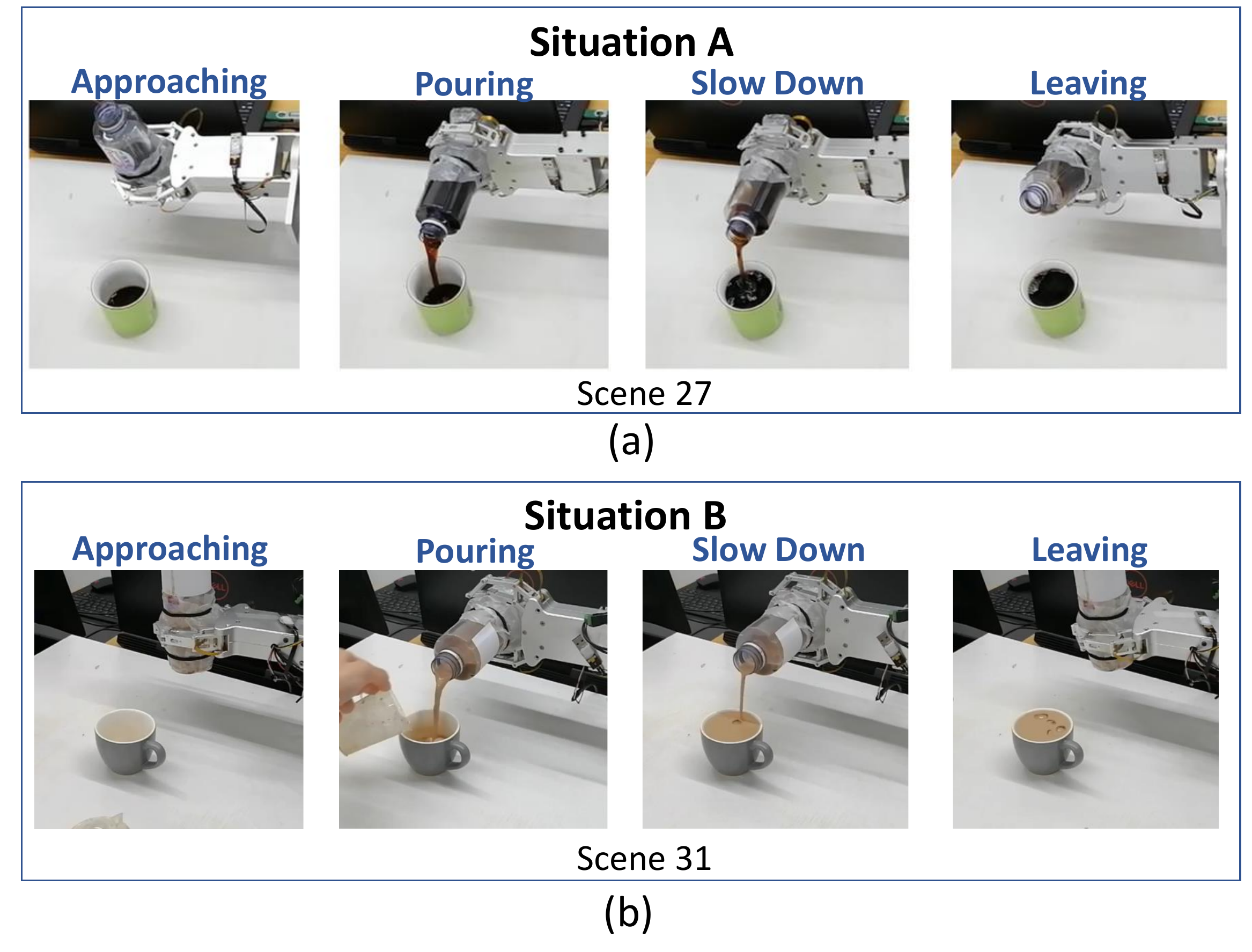}
	\captionsetup{font=footnotesize,labelsep=period}
	\caption{Two cases for verification of the second set of experiments. (a) A successful trial for Situation A, where the target container is not empty initially, the state transition of the pouring task may skip four states at the beginning of the pouring process. (b) A successful trial for Situation B, where additional ice cubes are added to the target container during the pouring process.} 
	\label{fig:Experiment-Adaptability.pdf}
\end{figure}

Table~\ref{tab:Manipulability11} lists eight testing cases.
Cases C1--C4 represent \textit{Situation A} and the determined initial state $k$ for each case is displayed in the third column. Cases C5--C8 represent \textit{Situation B}, for which the determined states $k_1$ and $k_2$ are also provided. Two successful trials  for the verification of the second set of experiments are shown in Fig. \ref{fig:Experiment-Adaptability.pdf}.
The success rates (SR) of the proposed method are summarized in the fourth and last columns for \textit{Situation A} and \textit{Situation B}, respectively.

Results indicate that the success rate of the proposed method for \textit{Situation A} is high, which means that it can be adapted to this situation with ease.
By contrast, the interruption during the pouring process brings extra challenges to ensure consistency in \textit{Situation B} and more adaptive methods need to be developed in the future.

\subsection{Manipulability}

The third set of experiments targets  the evaluation of the manipulability of the proposed method. The logical graph can incorporate programmatic features to enable manipulability for pouring drink with various target volume to a target container. In this way, human operators can manipulate the logical graph without collecting new demonstration data and retraining the model from scratch. The general situation represents pouring drink into an empty target container and fill the target container to be 90\% full. Ten trials were conducted for ten different scenes where the  final state of pouring is different from the general situation. 

\begin{figure*}[tb]
	\centering
	\includegraphics[width = 0.95\hsize]{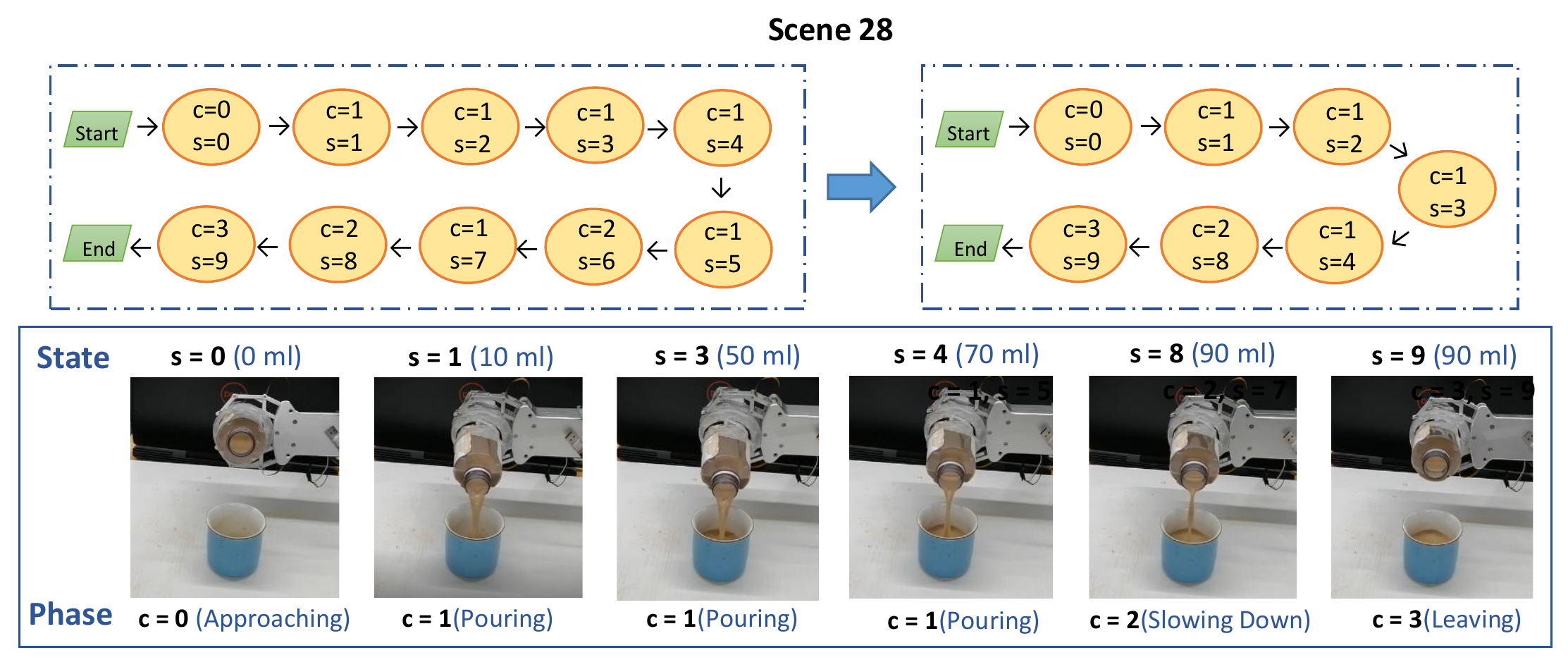}
	\captionsetup{font=footnotesize,labelsep=period}
	\caption{A successful experiment of drink pouring with a new goal in Scene 28 as an example for demonstrating the manipulability of the proposed method.}
	\label{fig:manipulated}
\end{figure*}

In most of the scenarios, we expect the robot to fill up the target container to be 90\% full. However, if we want the model to be manipulated for other goals of pouring task, for example, filling up the target container to be 70\% full, we can modify the logical graph by skipping several states before the ``slowing down" phase and the ``leaving" phase. An example is shown in Fig. \ref{fig:manipulated} to demonstrate the manipulation of the logical graph. For Scene 28, the original target is to pour drink with a total amount of 150 $ml$  into the target container, while the new target is to pour drink with a total amount of 90 $ml$. Therefore, the manipulated logical graph should skip the $s=5,6,7$ during the drink pouring phase, i.e. after 70 $ml$ liquids ($s=4,c=1$) has been poured into the target container, the next state and phase should be $s=8,c=2$, instead of $s=5,c=1$ in the original logical graph.

\begin{table}[!htb]
	\centering
	\captionsetup{font=footnotesize,labelsep=period}
	\caption{\textsc{Experimental Results for the Validation of Manipulability.}}
	\label{tab:Manipulability1}
	\begin{tabular}{c|c|c|c}
		\hline
		\textbf{Scene}  & \textbf{Target Vol. (ml)} & \textbf{Maximum Vol. (ml)} & \textbf{Success Rate} \\\hline	
		28	    & 90 & 150 & 0.50 \\	
		32	   & 70 & 130 & 0.60 \\			
		32	  & 110 & 130 & 0.90  \\			
		34	   & 90 & 150 & 0.80 \\			
		34	  & 130 & 150 & 0.90  \\			
		\hline
	\end{tabular}
\end{table}

Detailed experimental results for validation are shown in Table~\ref{tab:Manipulability1}. The overall success rate is 0.84 for all the experimental trials. Without EHIL, the baseline method cannot perform these types of drink pouring tasks, unless new demonstration data is collected and the model is retrained. 
With EHIL for performing drink pouring task with different initial conditions and final targets, it is also noticed that the success rate is not high for some experiments.  The success rate has a close relationship with the phase and state estimation accuracy, which needs to be further enhanced. To improve the perception of the liquid amount for state estimation, for example, an RGB-D camera can be used instead of the only RGB camera used in this work.

\subsection{Explainability}
\label{sec:explainability}

\begin{figure}[tb]
	\centering
	\includegraphics[width = 1 \hsize]{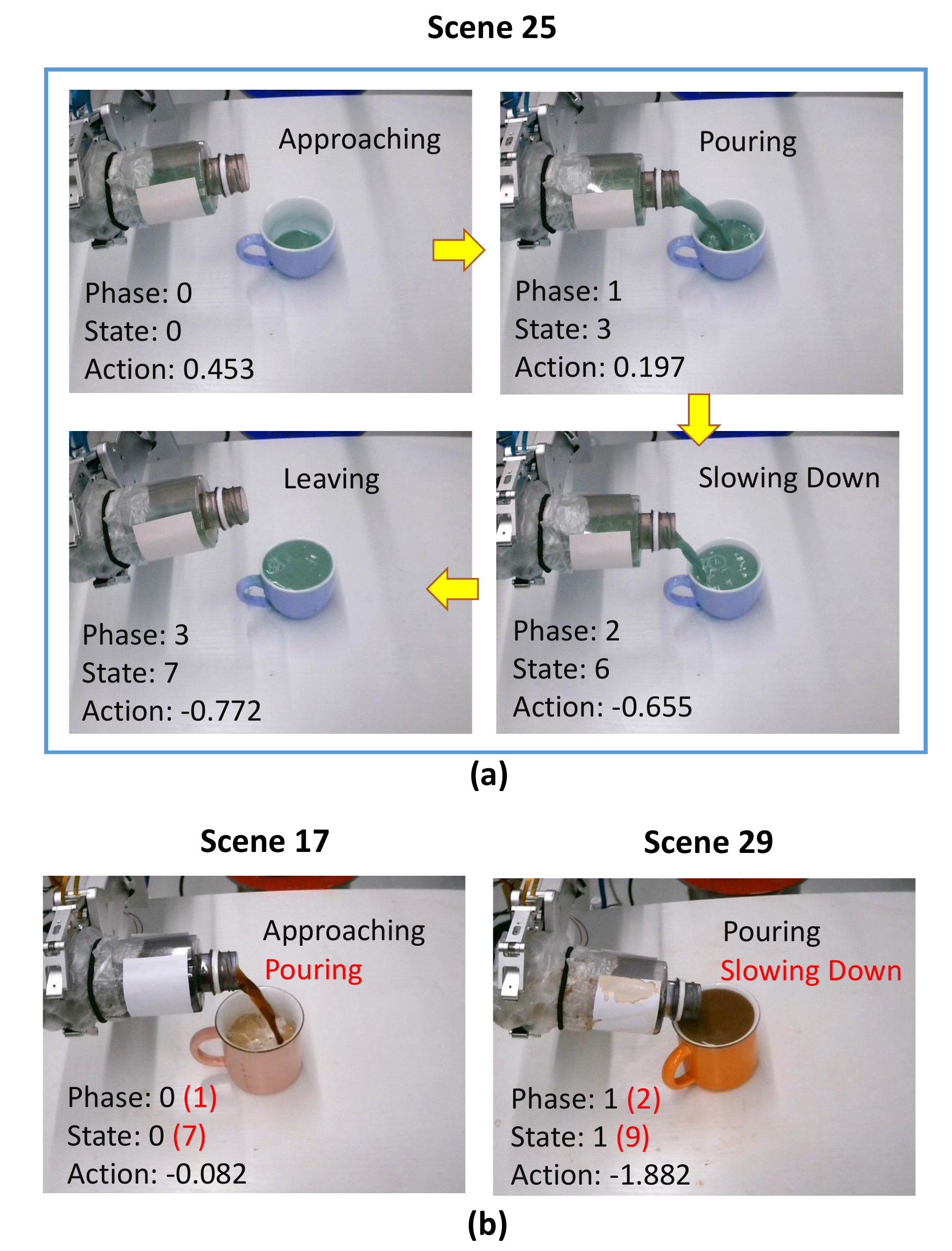}
	\captionsetup{font=footnotesize,labelsep=period}
	\caption{Comparison of phases and states determined by the pretrained models (in the black text) with corresponding labels (in the red text). (a) The determined values match the labels (not shown separately) in Scene 25, by which we predict that the task can be accomplished by the current models. (b) In Scene 17 (with milk in the target container initially)  and Scene 29 (with an adjusted camera view), the determined values and labels do not match at some frames.}
	\label{fig:xai}
\end{figure}

Humans can understand the decision of the model with trust, since the proposed algorithm generates not only action values but also high-level decision-making features.
The network can be further improved with clear direction after failure analysis while the black-box effects caused by the deep neural network model can be eliminated to some extent. 
The logical graph can also incorporate programmatic features to enable pouring drink with different total amounts of drink to the target container with different initial amount of liquids inside. 
This paves a way for the development of explainable intelligent robots, which can benefit the applications in unstructured environments or complex tasks that involve human-robot interaction. 
The explainability for enhancing user' trust by result prediction and post-hoc analysis as well as failure tracing is illustrated as follows.

\begin{figure*}[tb]
\centering
\includegraphics[width = 1\hsize]{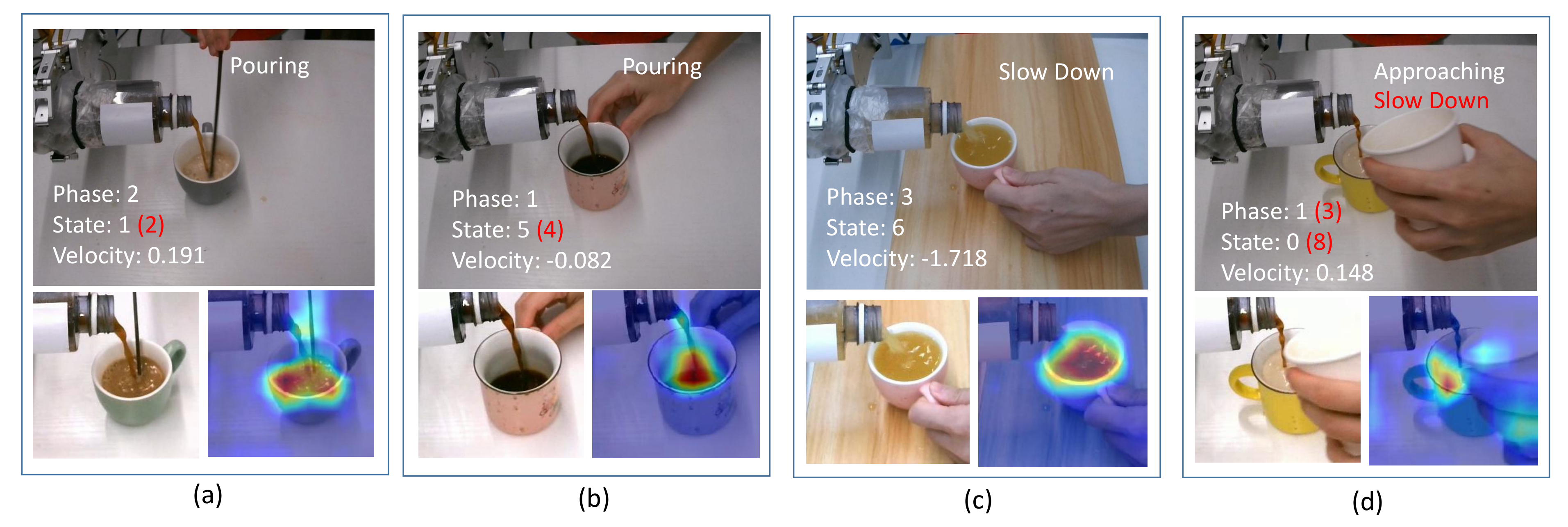}
\captionsetup{font=footnotesize,labelsep=period}
\caption{  Visualization of the post-hoc explanation for  (a) the scenario of coffee making with a stick stirring the coffee after adding milk and sugar, (b) the scenario of a human hand suddenly appearing in the camera view. (c) The scenario of pushing the target container to adjust its position slightly, while a new plate is placed under the target container. (d) The scenario of adding milk to the target container with a big container which can occlude the camera view during the black coffee pouring process.} 
\label{fig:XAI-Final}
\end{figure*}

\subsubsection{Enhancing User's Trust by Result Prediction}

The fourth set of experiments is to show how the explainability of the EHIL method in terms of the logical graph can help users predict potential failures and trust the decision-making of the robot prior to task execution, since the logical graphs are conducive to human interpretation.
Before applying the method to a new pouring task, we can teleoperate the robot to demonstrate the task as we do in the collection of training data.  
With the recorded data, we can use models $F_1(.)$ and $F_2(.)$ from the first and second hierarchies to generate the logical graph and check whether the phase and state transitions in the logical graph are reasonable or not by comparing it with the pre-defined desirable logical graph. These would assist humans to infer whether the intended actions of the robot will be correct or not. If the task is predicted to fail, the models have to be fine-tuned with the new demonstration data. Suppose that $\hat{c}(t)$ and $\hat{s}(t)$ are the predicted phase and state determined by the EHIL model at time step $t$, respectively, while $c(t)$ and $s(t)$ are the expected phase and state determined by human operators, respectively. The following three criteria are used to determine whether the task is predicted to fail or not:

\begin{enumerate}
\item The phase or state transition process does not follow the pre-defined sequential logics, for example,  $\hat{c}(t+1)$ $<$ $c(t)$ or $\hat{s}(t+1)$ $<$ $s(t)$ occurs, since liquid cannot return to the source container once it flows outside the source container (pouring is a non-reversible process);
\item The predicted phase transition process does not strictly follows the context of ``approaching", ``pouring", ``slowing down", and ``leaving" 
successively, since it is unreasonable to skip one of two phases during the pouring task; % while the average estimation probability higher than 0.90 during each phase,
\item The predicted states between two consecutive frames are larger than 3, i.e., $\hat{s}(t+1)$ - $\hat{s}(t)$ $>$ 3 (when $\hat{s}(t+1)$  $<N$), since it is impossible to add a large amount of liquids to the target container within $\delta{t} = \frac{1}{f}$, where $f$ represents the control frequency.
\end{enumerate}

Here, ten scenes (Scenes 16--20 and 25--29) are selected to evaluate the success rate of such a prediction using the models that are trained with the data collected in Scenes 1--20.
To add disturbances to the tests, milk is added into the target container initially in Scene 17 and the camera view is slightly changed in Scene 29 from where it is in collecting the training data.

Based on the three criteria mentioned above, failure prediction can be conducted based on the generated logical graph.	Fig.~\ref{fig:xai} shows some prediction results, where the models are competent for Scene 25 since the determined phases and states match their corresponding labels, as shown in Fig.~\ref{fig:xai}a, but not for Scenes 17 and 29, as shown in Fig.~\ref{fig:xai}b.
In practice, two tasks (Scenes 17 and 29) that are predicted to fail indeed fail in the end. Among the other eight tasks predicted to be successful, seven tasks succeed and the task in Scene 20 fails. 
That is to say, 90\% of the predictions are correct. 
In this way, users can predict whether the models on hand are suited for online deployment or not.
	
\begin{figure*}[tb]
\centering
\includegraphics[width = 1 \hsize]{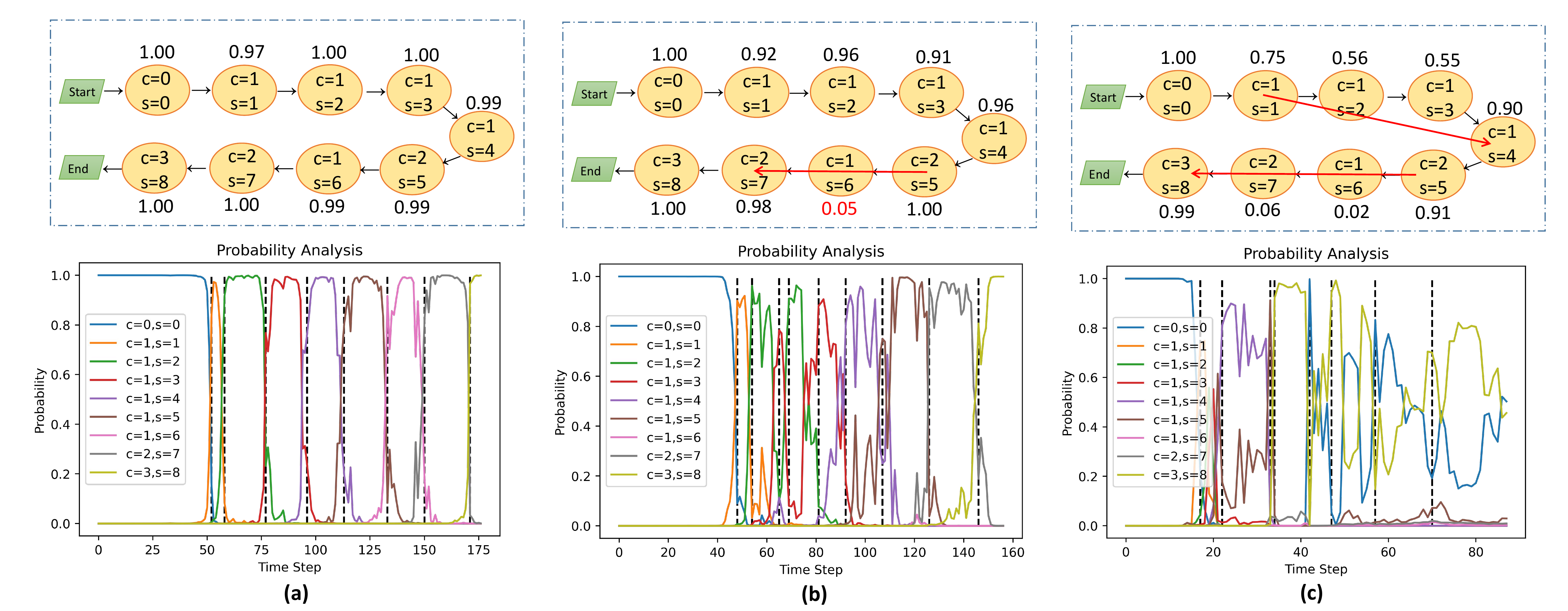}
\captionsetup{font=footnotesize,labelsep=period}
\caption{Tracing the causes of failure based on the logical graph with the probability of state determination. Red arrows represent unexpected state transitions, which skip some true states due to their low estimated probabilities. (a) Scene 9. (b) Scene 10. (c) Scene 32.}
\label{fig:prob}
\end{figure*}

\subsubsection{Enhancing User's Trust via Post-Hoc Explanation}

The fifth set of experiments is to check if the proposed method can be applied to the case where users accidentally disturb the pouring process, since service robots may have to face such unexpected disturbances in practical applications.

Fig.~\ref{fig:XAI-Final}a shows a typical coffee-making process with a stick stirring the coffee after adding milk and sugar, while Fig.~\ref{fig:XAI-Final}b shows the case where a human hand suddenly approaches the target container.
By selecting the region of interest for analysis, the brightness and the contrast of the cropped image are adjusted and Gradient-Weighted Class Activation Mapping (Grad-CAM) \cite{selvaraju2017grad,zhang2021surgical} is used to visualize how the deep neural network determines phases and states. 
The class-specific gradient information can be conveyed to the final convolutional layer of the deep neural network. The critical parts that contribute the most to the phase and state determination can be visualized by a coarse localization map with important regions highlighted.

In both cases, all phase determinations are correct, but some state estimations are not correct. 
In Fig.~\ref{fig:XAI-Final}a, this is because the color of the coffee is changed significantly after too much milk is added. In Fig.~\ref{fig:XAI-Final}b, the approaching of human may affect the lighting condition, which leads to the inaccuracy of state estimation.
However, the overall pouring process can still be accomplished and it seems that the incorrect state estimation does not influence the final performance.

As shown in Fig.~\ref{fig:XAI-Final}a and b, the red color highlights the most discriminative image region used by the model to determine phases and it appears that the region depends on the shape of the target container and the liquid surface in it.
Therefore, either the stick, milk, and sugar in coffee making or the human hand in the other case could not influence the region significantly, and users can trust the robot's decision-making based on the proposed method in facing these disturbances.

The plate used in the scenario shown in Fig. \ref{fig:XAI-Final}c is not included in the model training process as well, but the drink pouring process will not be influenced, since the model can neglect the background and pay attention to the liquid status to generate the prediction results.

Fig. \ref{fig:XAI-Final}d represents the scenario of adding milk to the target container with a big container which can occlude the camera view during the black coffee pouring process.  The phase predicted by the neural network was ``approaching'', while the real phase should be ``slowing down''. The expected state should be $s=0$, while the predicted state generated by the network is $\hat{s}=8$, which has significant deviation. This scenario is predicted to fail, since it meets the third criteria of failure prediction. The visualization result demonstrates the occlusion caused by the big container disables the correct prediction of the phase and state.

\subsubsection{Tracing out Failures}
\label{sec:causes}

The sixth set of experiments further shows how the causes of failure can be traced out by using the logical graph. 
For this we chose three scenes (Scenes 9, 10, and 32) for evaluation, among which Scene 9 has been used in the model training, Scene 10 has a changed lighting condition, and Scene 32 represents an unseen scene.

Fig.~\ref{fig:prob} shows the logical graphs with the probability for each determined phase and state generated by models $F_1(.)$ and $F_2(.)$.
In the experiments, the robot succeeded in pouring drink in Scenes 9 and 10 but failed in Scene 32.
Since Scene 9 has been used in training models, it is not surprising that all the state estimation probabilities of the correct state are high while the state transition processes are clear and every state transition happens as expected, as shown in Fig.~\ref{fig:prob}a.
With the change of lighting, the state probabilities and transitions in Scene 10 have some exceptions, as shown in Fig.~\ref{fig:prob}b, but the situation is not too severe to fail the whole task while the critical phases are determined correctly.
In the logical graph for Scene 32 shown in Fig.~\ref{fig:prob}c, however, many key states are mistakenly skipped, which leads to the failure of the task. This may be caused by the varying lighting conditions and the properties of the latte used for experiments, which results in the ill performance of the model during the test at that specific period a time.

\section{Discussions}
\subsection{Model Performance}
The EHIL proposed in this paper requires collecting the robot demonstration data via teleoperation. After that, the hierarchical neural network model is constructed for phase and state determination, which is trained in a classification manner. The low-level action determination hierarchy is trained in regression manner, while safety constraints can be added to ensure safety.  The knowledge learned from the first hierarchy is known as the common sense for all types of drink pouring task, which is invariant to the target and source containers. This follows the way of how humans learn and generalize strategies for  everyday tasks. In this way, the robot can infer high-level concepts and can easily transfer knowledge and perform various tasks among different scenarios. They are not simply replicating the operators' behaviors but learning how and why to perform the tasks in the meantime. 

The second hierarchy can be retrained when the total number of states is changed. The number of the states depends on the tolerable error for the task. More states can improve the accuracy of pouring but may need more data for model training. The trade off between the state number and the amount of data required to be collected should be considered, which depends on the application scenarios. For example, the amount of liquids poured by service robots for assisting people's daily life do not need to be very precise, so the tolerable error can be set to be 10 ml. In this case, the resolution of each state is set to be $k$ =  20 ml, while  $N = \frac{100}{20} = 5$,  the total number of the states for pouring drink to a target container with maximum capacity of 100 ml is $6$. As for pouring liquids in wet laboratory, the precision for controlling the liquid amount should be higher. Assume that the tolerable error is 5 ml, $N = \frac{100}{10}  = 10$, the total number of states required for the same target container is 11. 

The models trained for the first and second hierarchies are general to be adapted to multiple robotic platforms but the implementation of the third hierarchy requires demonstration data collected from the specific robotic platform. Therefore, the proposed method can be transferred to different robotic platforms by retraining the weights of hierarchy 3, while the models of hierarchies 1 and 2 only need to be fine-tuned. In this way, the hierarchical imitation learning framework can be implemented on multiple platforms with ease.

As for the setting of the neural network structure of  EHIL, it is not fully optimized. Neural architecture search (NAS) can be used to automate the selection of the optimal parameters and structure to optimize the neural network model. However, this is out-of-scope for detailed discussion, since we focus on leveraging the advantages of deep neural network models and classic symbolic AI (like logical graph) to empower the robots with the capability of providing  explanations of their behaviors and enhancing adaptability.

\subsection{Applicability to Other Applications}

For more dynamic and complex scenarios, the robot should be incorporated with the ability of finding the target container automatically. This is not the focus of this paper and we assume that some out-of-the-box-solutions can be used to reach this target \cite{schenck2017visual}.

The experimental setup in this paper includes a low-cost RGB camera for data acquisition as sensory input, which can enable the ease of deployment of the proposed method in different environments. 	 However, the use of an RGB camera for data acquisition is not a must for the implementation of the proposed method. Depth camera (RGB-D camera) or stereo camera can be used to provide more comprehensive information while the encoders for capturing the joint angle information of the robot can be used as the input as well. Force or tactile information can also be used, provided that the corresponding sensors are mounted on the wrist of the robotic arm. Different sensor information can be incorporated to the framework to formulate a multi-sensor fusion based imitation learning framework.

%\textcolor{red}{Moreover, the actions that we defined in this paper are the velocities of the end effector. For other robotic platforms using torque control, the generated action can be replaced by joint torque value, which can be mapped to end-effector pose value for online deployment. }

%\textcolor{red}{Last but not least, though our experiments focus on the robotic pouring task, the proposed method can be extended to, for example, washing dishes, clearing and setting a table by manipulating a various number of dishes or bowls, building a pyramid of cups etc. }

Moreover, the proposed method can be adapted to different robotic platforms with different control modes. The action that we defined in this paper is the velocity of the end effector. For other robotic platforms which require torque control, the control commands can be mapped from the end-effector's pose and velocity for online deployment.

Last but not least, though our experiments focus on the robotic pouring task, the proposed method can be extended to a series of sequential tasks, for example, opening a door, washing dishes, clearing and setting a table by manipulating a various number of dishes or bowls, building a pyramid of cups, and etc. The key feature of these tasks is that the phase and state can be clearly defined. Take the door opening task as an example, the predicted phase transition process should strictly follow the context of ``approaching the door" ($c=0$), ``grasping the door handle" ($c=1$), ``opening the door" ($c=2$), and ``leaving the door handle" ($c=3$) successively. During the phase of ``opening the door", different states can be used to determine the opening angle of the door. For example, $s=0$ represents that the door is completely closed. Suppose that the maximum angle of door opening is $90^{\circ}$, while the resolution is set to $k=10^{\circ}$. Therefore, $N = \frac{90^{\circ}}{10^{\circ}} = 9$, the total number of state is 10, i.e. $s=10$ represents that the door is completely opened. The action value can be determined as the 6-D end-effector pose of the robot, which is the output of hierarchy 3. In this way, the EHIL method can be extended to this new task, and similar concepts can be used for imitation learning of many other tasks for the development of intelligent robots.
	
\section{Conclusions and Future Work}
In this paper, we propose a method to enable robots to perform drink pouring tasks automatically via hierarchical imitation learning. 
To make the decision-making process explainable to users, we construct a logical graph to show the phase and state transitions during the execution of a pouring task, which  can enhance user's trust in the automatic robotic drink pouring process and provide a guide to trace the causes of failure in the task. With the proposed method, the robot can explain its decisions and justify its actions to a human. 
We also verify that the proposed method can be applied to some unseen scenes and has adaptability in some drink pouring scenarios that have not been demonstrated before.
%of pouring drink
	
As for the future work, semi-supervised or unsupervised learning techniques can be employed such that the manual annotation process for supervised learning can be avoided. 
Also, since it is much easier to collect human demonstration data than robot demonstration through teleoperation, it is worthwhile to explore visual imitation learning by watching human demonstration videos and transferring knowledge and control policies learned from human to different robotic platforms. 
Moreover, the adaptability of the proposed models can be further enhanced by enabling robots to handle new types of drinks and containers with complex shapes.
%The automatic grasping of source container and drink pouring to multiple target containers can be explored, which can enable the robot to be an intelligent server in pubs or restaurants and interact with humans with explainable behaviors.		

\section*{Acknowledgements} 
The authors would like to thank K. Chen, W.-W. Lee, Y. Liu, and K. Xiong for facility support. %Qiang Li would like to acknowledge the support by the Deutsche Forschungsgemeinschaft (DFG, German Research Foundation) – Projektnummer (No.410916101).
	
%The authors would like to thank Ke Chen, Wang-Wei Lee, Yuezhang Liu, and Kun Xiong for preparing the customized robotic gripper and other parts used in the experiments.  

\bibliographystyle{IEEEtran}
% argument is your BibTeX string definitions and bibliography database(s)
\bibliography{references}

\end{document}